\documentclass[journal]{IEEEtran}
\usepackage{cite}
\ifCLASSINFOpdf
\else
\fi

\usepackage{enumitem}

\usepackage{booktabs}       % professional-quality tables
\usepackage{color}
\usepackage{eucal}
\usepackage{amssymb,amsmath,amsthm}
\usepackage{dsfont}
\usepackage{multirow}
\usepackage{graphicx}
\usepackage{subfig}
\newcommand{\alg}{SNACS}
\newcommand{\norm}[1]{\left\lVert#1\right\rVert}
\usepackage[ruled,vlined]{algorithm2e}

\newtheorem{theorem}{Theorem}

\def\bbE{\mathbb{E}}
\def\bx{\mathbf{x}} \def\by{\mathbf{y}}\def\bz{\mathbf{z}}
\def\bX{\mathbf{X}}\def\bY{\mathbf{Y}}\def\bZ{\mathbf{Z}}
\def\diy{\displaystyle}

\begin{document}

%
% paper title
% Titles are generally capitalized except for words such as a, an, and, as,
% at, but, by, for, in, nor, of, on, or, the, to and up, which are usually
% not capitalized unless they are the first or last word of the title.
% Linebreaks \\ can be used within to get better formatting as desired.
% Do not put math or special symbols in the title.
\title{Slimming Neural Networks Using Adaptive Connectivity Scores}
%
%
% author names and IEEE memberships
% note positions of commas and nonbreaking spaces ( ~ ) LaTeX will not break
% a structure at a ~ so this keeps an author's name from being broken across
% two lines.
% use \thanks{} to gain access to the first footnote area
% a separate \thanks must be used for each paragraph as LaTeX2e's \thanks
% was not built to handle multiple paragraphs
%

\author{Madan~Ravi~Ganesh,
        Dawsin~Blanchard,
        Jason~J.~Corso,~\IEEEmembership{Senior Member,~IEEE,}
        and~Salimeh~Yasaei~Sekeh,~\IEEEmembership{Member,~IEEE}% <-this % stops a space
\thanks{Madan Ravi Ganesh is with the Department
of Electrical and Computer Engineering, University of Michigan, Ann Arbor,
MI, USA.}% <-this % stops a space
\thanks{Dawsin Blanchard and Salimeh Yasaei Sekeh are with the School of Computing and Information Science, University of Maine, ME, USA.}% <-this % stops a space
\thanks{Jason J. Corso was previously with the University of Michigan and is currently with the Stevens Institute for Artificial Intelligence, Stevens Institute of Technology, NJ, USA}% <-this % stops a space
}

% note the % following the last \IEEEmembership and also \thanks - 
% these prevent an unwanted space from occurring between the last author name
% and the end of the author line. i.e., if you had this:
% 
% \author{....lastname \thanks{...} \thanks{...} }
%                     ^------------^------------^----Do not want these spaces!
%
% a space would be appended to the last name and could cause every name on that
% line to be shifted left slightly. This is one of those "LaTeX things". For
% instance, "\textbf{A} \textbf{B}" will typeset as "A B" not "AB". To get
% "AB" then you have to do: "\textbf{A}\textbf{B}"
% \thanks is no different in this regard, so shield the last } of each \thanks
% that ends a line with a % and do not let a space in before the next \thanks.
% Spaces after \IEEEmembership other than the last one are OK (and needed) as
% you are supposed to have spaces between the names. For what it is worth,
% this is a minor point as most people would not even notice if the said evil
% space somehow managed to creep in.

% The paper headers
\ifCLASSOPTIONpeerreview 
\markboth{Journal of \LaTeX\ Class Files,~Vol.~14, No.~8, August~2015}%
{Shell \MakeLowercase{\textit{et al.}}: Bare Demo of IEEEtran.cls for IEEE Journals}
\fi
% The only time the second header will appear is for the odd numbered pages
% after the title page when using the twoside option.
% 
% *** Note that you probably will NOT want to include the author's ***
% *** name in the headers of peer review papers.                   ***
% You can use \ifCLASSOPTIONpeerreview for conditional compilation here if
% you desire.

% If you want to put a publisher's ID mark on the page you can do it like
% this:
%\IEEEpubid{0000--0000/00\$00.00~\copyright~2015 IEEE}
% Remember, if you use this you must call \IEEEpubidadjcol in the second
% column for its text to clear the IEEEpubid mark.

% use for special paper notices
%\IEEEspecialpapernotice{(Invited Paper)}

% make the title area
\maketitle

% As a general rule, do not put math, special symbols or citations
% in the abstract or keywords.
\begin{abstract}
In general, deep neural network (DNN) pruning methods fall into two categories: 1) Weight-based deterministic constraints, and 2) Probabilistic frameworks.
While each approach has its merits and limitations there are a set of common practical issues such as, trial-and-error to analyze sensitivity and hyper-parameters to prune DNNs, which plague them both.
In this work, we propose a new single-shot, fully automated pruning algorithm called \textit{Slimming Neural networks using Adaptive Connectivity Scores} (\alg{}).
Our proposed approach combines a probabilistic pruning framework with constraints on the underlying weight matrices, via a novel connectivity measure, at multiple levels to capitalize on the strengths of both approaches while solving their deficiencies.
In \alg{}, we propose a fast hash-based estimator of \textit{Adaptive Conditional Mutual Information} (ACMI), that uses a weight-based scaling criterion, to evaluate the connectivity between filters and  prune unimportant ones.
To automatically determine the limit up to which a layer can be pruned, we propose a set of operating constraints that jointly define the upper pruning percentage limits across all the layers in a deep network.
Finally, we define a novel \textit{sensitivity} criterion for filters that measures the strength of their contributions to the succeeding layer and highlights critical filters that need to be completely protected from pruning.
Through our experimental validation we show that \alg{} is faster by over $17$x the nearest comparable method and is the state of the art single-shot pruning method across three standard Dataset-DNN pruning benchmarks: CIFAR10-VGG16, CIFAR10-ResNet56 and ILSVRC2012-ResNet50.

\end{abstract}

% Note that keywords are not normally used for peerreview papers.

% For peer review papers, you can put extra information on the cover
% page as needed:
% \ifCLASSOPTIONpeerreview
% \begin{center} \bfseries EDICS Category: 3-BBND \end{center}
% \fi
%
% For peerreview papers, this IEEEtran command inserts a page break and
% creates the second title. It will be ignored for other modes.
\IEEEpeerreviewmaketitle

\section{Introduction}
% Problem Domain
Critical real-world applications like autonomous vehicle navigation~\cite{bechtel2018deeppicar,fridman2017autonomous} and simultaneous machine translation~\cite{gu2016learning,jia2019direct} demand real-time response~\cite{lin2018architectural} without any compromise in performance.
Proposed solutions in these application domains are implicitly bound to constraints brought forward by restricted space and memory availability on custom hardware implementations.
These factors are at odds with the general research design goal of high performance in deep neural networks (DNN), which is often achieved by increasing the overall size and capacity of the DNN.
The trade-off between these constraints has brought increased attention to the field of DNN pruning~\cite{gale2019state,liu2018rethinking}, whose main objective is to maintain an adequate level of performance, often within a few percent of the original DNN, while only using a fraction of its memory or FLOPs.  
Of course, the adequacy of any level of performance depends on the specific application, but the general goal nevertheless remains critical.

% How is it commonly approached, why it isn't sufficient
There are two main approaches to pruning: 1) deterministic constraints on weight matrices~\cite{wen2016learning,luo2017thinet,han2015learning} and 2) probabilistic frameworks~\cite{ganesh2020mint,dai2018compressing,luo2017entropy}.
Methods based on deterministic constraints on weight matrices are straightforward to implement and do leverage the underlying structure of the weight matrices, but they often do not account for the downstream impact of pruning filters.
On the other hand, probabilistic frameworks focus on reducing the redundancy between layers using information theoretic measures or variational bayesian inference but are not fast or efficient at modelling the sensitivity of filters at an individual level. 
% On the other hand, probabilistic frameworks focus on the connectivity between filters and their downstream impact but are not fast or efficient at modelling the sensitivity of filters at an individual level. 
In a sense, the two types of method are converses: one's weakness is remedied by the other.
Yet, to the best of our knowledge, there has been no recent work that combines both approaches and improves upon them.
Further, there are many unresolved practical issues among both approaches, e.g., the labor intensive process of analyzing the sensitivity of different layers to pruning or imposing an upper limit on pruning percentage for each layer and the amount of resources and time spent in iteratively pruning DNNs.

% Introduce our method and mention how it overcomes the prev. issues (Write it up similar to assertion of contribution and then how it helps avoid previous issues)
To that end, we are able to unify the benefits of both methods while mitigating their respective drawbacks:
we propose \textit{Slimming Neural networks using Adaptive Connectivity Scores} (\alg{}) as a hybrid single-shot pruning approach.
In \alg{}, we introduce the \textit{Adaptive Conditional Mutual Information} (ACMI) measure, which incorporates weights as a scaling function within the framework of conditional mutual information~\cite{Yurietal2016,TC}.
The ACMI measure evaluates the connectivity between pairs of filters across adjacent layers and prunes unimportant filters.
In this work, we  explore weight and activation-based scaling functions. 
% explore a multitude of scaling functions and highlight the improvement offered by weight-based functions.

\begin{figure*}[ht!]
    \centering
    \includegraphics[width=2\columnwidth]{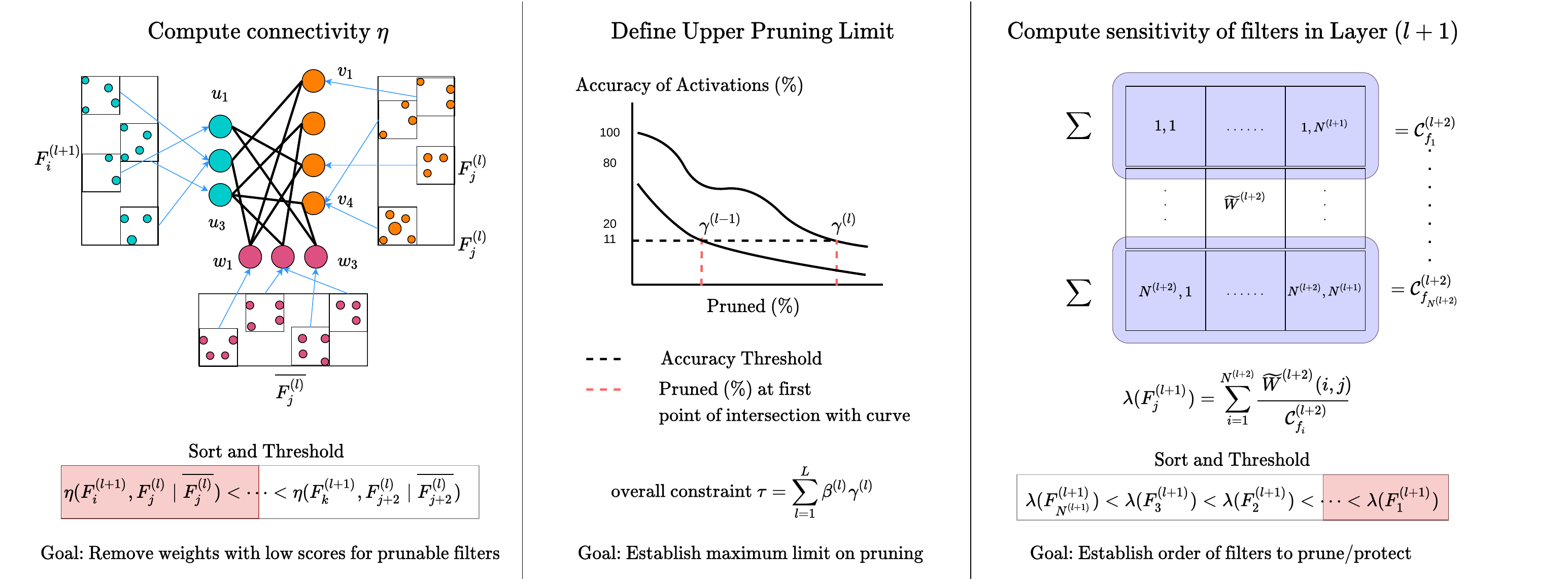}
    \caption{Illustration of the three major components of \alg{} that help prune connections between layer $l$ and $l+1$.
    First, we propose the hash-based ACMI estimator to compute the connectivity scores between filters in layer $l+1$ and all the filters in layer $l$.
    These connectivity scores are thresholded to obtain the set of filters that need to be pruned.
    Next, to protect the network from being irregularly/excessively pruned, we  use a custom set of operating constraints, based on the degradation of activation quality at various pruning levels, to decide on the upper pruning percentage limit for layer $l+1$.
    Finally, we compute the sensitivity of filters in $l+1$ as the sum of normalized weights between chosen filters in layer $l+1$ and all the filters in layer $l+2$. 
    We sort and threshold the sensitivity values to create a subset of sensitive filters which should be protected from pruning.
    Combining the information from all three components we prune layer $l+1$.}
    
    \label{fig:snacs}
\end{figure*}

To remove the manual effort involved in setting the upper pruning percentage limit of layers, we define a set of operating constraints to automatically evaluate them.
The constraints are based on the degradation in quality of activations at various levels of compression.
Additionally, we encapsulate the importance of a filter using our proposed \textit{Sensitivity} criterion, defined as the sum of a filter's contributions (normalized weights) to filters in the succeeding layer.
Using this measure, we curate a subset of relatively less sensitive filters that can be pruned based on their connectivity scores while we protect highly sensitive filters from any form of pruning. 
% By using \textit{Sensitivity} in conjunction with our set of constraints to define the upper pruning percentage limit of layers, we can boost the overall compression ($\%$) while maintaining test accuracy as well as automate the entire experimental pipeline.
We highlight all the main components of \alg{} in Fig.~\ref{fig:snacs}.

Overall, we summarize our contributions in this work below,
\begin{itemize}[topsep=0pt]
    \item We propose a hybrid single-shot pruning approach, \alg, which takes advantage of both a probabilistic pruning framework and simple weight-based constraints.
    \item In \alg{}, we propose the use of Adaptive Conditional Mutual Information (ACMI) as a way to measure the connectivity between filters and derive its hash-table-based implementation,
    \item In the interest of simplifying the process of defining upper pruning percentage limits of layers in a DNN, we propose a set of operating constraints to help automate their definition, and
    \item  We apply a custom notion of \textit{Sensitivity} in filters, using their contribution to succeeding layers, to prioritize the pruning of largely insensitive filters while protecting highly sensitive ones.
\end{itemize}
By incorporating our contributions within the \alg~framework, we improve the overall run-time of the pruning algorithm by upwards of $17\times$, increase the accuracy of the estimator, create an entirely automated pruning pipeline while offering state of the art performance in single-shot pruning of DNNs.
% highly competitive pruning algorithm using only a \textbf{single} prune-retrain step while existing state-of-the-art methods use a multiple prune-retrain steps.

\section{Related Works}
\label{sec:related_works}

In the following subsections, we discuss prior works in pruning and mutual information (MI) estimators, as well as methods at their intersection.
Among pruning approaches there are two broad categories: 1) methods that use a deterministic constraint on the weight matrices, and 2) methods that use a probabilistic framework to reduce the redundancy and maintain the flow of information between layers.
Within the first category of methods, there is a subset that enforces sparsity by modifying the objective function while the remaining directly apply constraints on the weight matrices.

\subsection{Deterministic Constraints on Weight Matrix}

\noindent \textbf{Direct Constraint on Weight Matrices}:
Some of the earliest works in pruning used the second-order relationship between the objective function and weights of the network to evaluate and remove unimportant values~\cite{lecun1990optimal,hassibi1993second}.
Since then, several advancements in the form of directly thresholding weights~\cite{han2015learning,guo2016dynamic} or using the $\vert \vert.\vert \vert_1$ constraint to define the importance of filters~\cite{DBLP:conf/iclr/0022KDSG17} have been proposed. 
A more recent subset of methods have adopted data-driven logic to derive the importance of filter weights. 
Two such methods are ThiNet~\cite{luo2017thinet} and NISP~\cite{yu2018nisp}, where the reconstruction of outcomes with the removal of weights is posed as a post-training objective.
By virtue of how direct constraints are placed on weight matrices, they often do not account for the downstream impact of pruning or are built on the assumption of a purely deterministic relationship between filters.
Instead, we use the combination of a weight-based scaling function and filter connectivity within a probabilistic framework to maintain the flow of information between layers and overcome these issues.

\noindent\textbf{Modification of Objective Function}:
Inducing sparsity in weight matrices by modifying the objective function involves imposing a strong constraint on how weights develop during training.
Constraints range from simple methods, such as single or multiple $l_n$ norms~\cite{liu2017learning} on channel outputs, simple patterned masks to regulate group sparsity~\cite{lebedev2016fast}, and optimization over group-lasso-based objective functions~\cite{wen2016learning,huang2018data}, to more complicated ideas like balancing individual vs. group sparsity constraints~\cite{he2017channel,yoon2017combined} and adding discrimination-aware losses at intermediate layers to enhance and easily identify important channels~\cite{zhuang2018discrimination}.

More recent methods combine the idea of modifying the objective function with more abstract concepts like meta-learning~\cite{li2020dhp}, where sparsity inducing regularizers are used to learn latent vectors that help decide on the weight values directly or 
% the enforcement of an extra constraint on the weight values to prune the network.
% Such methods include,~\cite{li2019oicsr}, in which a custom group-lasso objective is optimized while pruning is based on a squared dependency on weight values, and
GANs, where an adversarial pruned network generator optimizes a loss based on the features derived from the original network~\cite{lin2019towards}.
%% SHAVE: Can shave if necessary
To provide a controlled set up to study and compare the effects of pruning a network against its original counter-part, we avoid strong comparisons against methods that modify the objective function.
Apart from optimizing over a fundamentally different objective function, which are harder to optimize, these methods require multiple iterations of pruning and fine-tuning built-in to their setup, while we use only a single pruning and retraining step while targeting a simple objective function.

\subsection{Probabilistic Frameworks}
Pruning approaches that use a probabilistic frameworks can be divided into bayesian and non-bayesian methods.
Bayesian methods apply a variational bayesian inference perspective to pruning, with a focus on estimating the posterior distribution of weights using ELBO~\cite{zhao2019variational,louizos2017bayesian}. 
While they offer a theoretically sound perspective to pruning, they require a strong assumption on the prior distribution of weights which induces sparsity across the network. Further, their performances on large-scale datasets have more room to grow.

The non-bayesian approach to pruning focuses on using information theoretic measures, with minimal assumptions and widespread applicability when compared to the bayesian methods. 
These include, Luo and Wu~\cite{luo2017entropy}, in which entropy of activations is used as a measure of importance of a filter, VIBNet~\cite{dai2018compressing}, where the information bottleneck principle is used to minimize the redundancy between adjacent layers and MINT~\cite{ganesh2020mint}, in which geometric conditional MI is used to determine the dependencies between filter pairs in adjacent layers.
While they are adept at reducing redundancy and maintaining the flow of information between layers, they are slightly slow and inefficient at modelling the sensitivity of individual filters to pruning.
In \alg{}, we propose the use of ACMI which improves the speed of dependency computations for MI as well as the accuracy of the estimates.
Further, by highlighting sensitive filters that need to remain un-pruned and jointly defining the upper pruning percentage limit of layers we obtain additional gains when pruning a DNN.
% overcome some of the common practical issues in previous methods while improving the overall performance.

\subsection{Multivariate Dependency Measures}
Approaches for estimating multivariate dependencies using MI can be broadly classified into two categories: plugin and direct estimation.
Plugin estimators like Kernel Density Estimators (KDEs)~\cite{kraskov2004estimating}, KNN estimators~\cite{moon2017ensemble}, and others \cite{principe2000information,yang2000data} form the bulk of early works in computing multivariate dependency.
However, plugin estimators need to accurately estimate the probability density function of input variables.
This, when combined with their large run-time complexity, renders them highly un-scalable.
To overcome these issues, direct estimators for Renyi-entropy and MI~\cite{principe2000information,yang2000data}, and Henze-Penrose divergence measure~\cite{leonenko2008class} have been proposed.
% Kullback-Liebler~\cite{pal2010estimation}
They provide manageable run-time complexity while avoiding direct knowledge of the density function.
Crucial to the functioning of many direct estimation methods is the use of graph-theoretic ideas, such as the Nearest Neighbour Ratios~\cite{noshad2017direct}, which uses the $k$-NN graph to estimate MI, and the minimum spanning tree used to estimate the GMI~\cite{yasaei2019geometric}. These graph-based approaches help make the evaluation of MI computationally tractable.
While most methods fall into either plugin or direct categories, recent work has focused on the development of a hybrid approach~\cite{Mortezaetal2019}.
This approach combines the fast run-time implementation of hash-tables with an error convergence rate akin to plugin methods, thus merging the advantages of both the estimation approaches.

% The accurate estimation of multivariate dependency in high-dimensional settings is a hard task. 
% The first in this line of work involved Shannon Mutual Information which was succeeded by a number of plug-in estimators including Kernel Density Estimators~\cite{kraskov2004estimating} and KNN estimators~\cite{moon2017ensemble}.
% However, their dependence on density estimates and large runtime complexity means they are not suitable for large scale applications including neural networks.
% Faster plug-in methods based on graph theory and nearest neighbour ratios~\cite{noshad2017direct} were proposed as an alternative.
% More solutions that use statistics like KL divergence like ~\cite{leonenko2008class} or Renyi-$\alpha$~\cite{gao2015efficient} were proposed to help bypass density estimation fully.
% Instead, in this work, we focus on a conditional GMI estimator, similar to \cite{yasaei2019geometric}, which bypasses the difficult task of density estimation and is non-parametric and scalable.

\section{Algorithm and Component Description}
\label{sec:detailed_design}

In the following subsections, we outline \alg's algorithm.
This is followed by details of the ACMI measure and the set of constraints that automatically define the upper pruning percentage limits of layers in a DNN.
The final sub-section explains our notion of sensitivity, which identifies and protects important filters from being pruned.

\begin{figure*}[ht!]
    \centering
    \includegraphics[width=\textwidth]{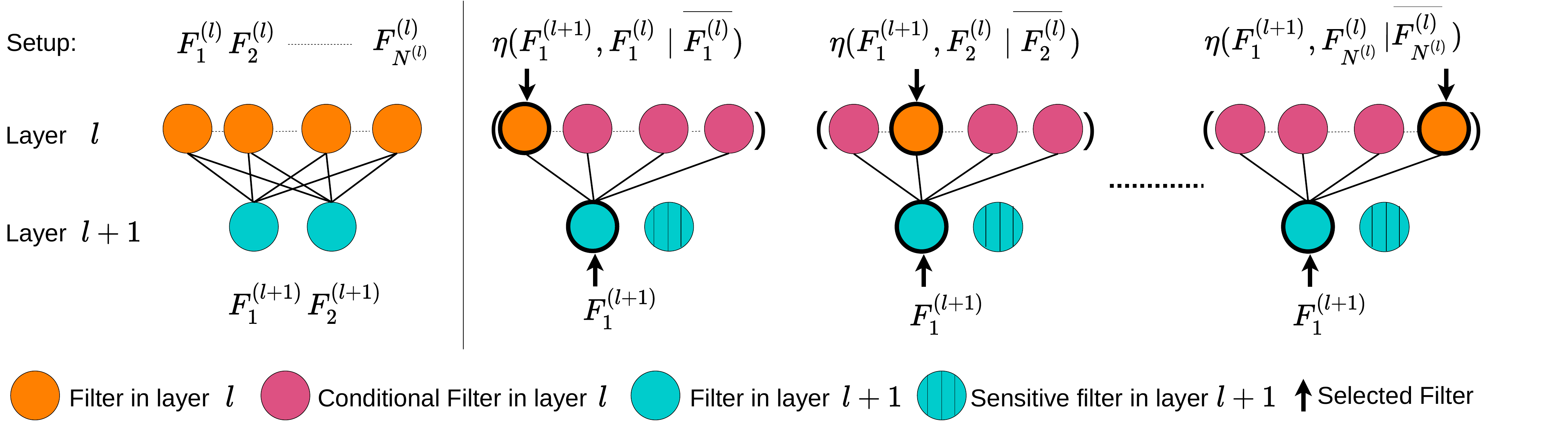}
    \caption{Example of computing ACMI, $\eta()$, between activations of filters in layers $l+1$ and $l$. In each $\eta()$ computation, the arrows indicate the filters between which we compute the connectivity score while taking into consideration the activations from the remaining filters in layer $l$. These steps are repeated for every possible pair of filters except for highly sensitive filters in layer $l+1$, where $\eta$ need not be computed since their connections (lines between filters) are not pruned.}
    \label{fig:snacs_dependency}
\end{figure*}

\subsection{Notation}
\label{sec:notations}
We assume that a given DNN has a total of $L$ layers where,
\begin{itemize}
    \item \textsc{Sensitive filters()} : Function that returns the indices of a subset of filters that need to be protected from pruning, computed using sensitivity (Section~\ref{sec:evaluating_sensitivity_of_filters}). 
    \item $F^{(l+1)}_i$ : Activations from the selected filter $i$ in layer $l+1$.
    \item $N^{(l+1)}$ : Total number of filters in layer $l+1$.
    \item  $S_{F^{(l+1)}_i}$ : The set of filter indices whose values are pruned from the weight vector. 
    \item $\eta{()}$ : Connectivity score between two filters computed using ACMI (Section~\ref{sec:weighted_hash_based_cmi_estimator}).
    \item $\overline{F^{(l)}_j}$: The set of all filters excluding $F^{(l)}_j$ in layer $l$.
    \item $\delta$: Threshold on connectivity scores to ensure only strong connections are retained.
    \item $\gamma^{(l+1)}$ : Upper limit on pruning percentage for layer $l+1$ defined using the constraints in Section~\ref{sec:algorithm_defining_upper_pruning_limit_of_a_layer}.
\end{itemize}

\subsection{Algorithm}
\label{sec:algorithm}

The overall goal of our algorithm is to find the set of filters that contribute minimally to the flow of information between layers and prune their values from the weight matrix.
We apply \alg{} between every pair of adjacent layers in a pre-trained DNN where,
\begin{itemize}
\item We identify a subset of sensitive filters in layer $l+1$ that need to be protected from pruning and iterate over the remaining insensitive filters in layer $l+1$.
% $S_{F^{(l+1)}_i}$ is initialized with the indices of these subset of filters.
\item To measure the connectivity score, $\eta$, between filters in layers $l$ and $l+1$ we apply our proposed hash-based ACMI estimator on the activations from each set of filters. 
An example of this is shown in Fig.~\ref{fig:snacs_dependency}.
The connectivity score evaluates the strength of the relationship between two filters in the context of contributions from all the remaining filters in layer $l$.
\item If the connectivity score is lower than a threshold level $\delta$, and the number of pruned filters do not exceed the pre-determined upper limit, denoted by $\gamma^{(l+1)}$, we add the index of the filter to $S_{F^{(l+1)}_i}$.
The weights for retained and protected filters/neurons are untouched while the weights for the entire kernel/elements are zeroed out for pruned filters/neurons. 
\end{itemize}
In the practical implementation of Alg.~\ref{alg:SNACS} the value of $\delta$ is determined by thresholding $\eta$ values from a chosen layer so as to remove sufficient weights and match the predetermined $\gamma^{l}$.
Once we prune the filters that contribute the least across all the layers of the DNN, we proceed to re-training the network using a setup that mirrors the training phase of the pre-trained DNN.
Across Alg.~\ref{alg:SNACS}, we note that \alg{} does not contain a continual feedback loop to update weights when pruning. 
Instead, we take only a single retraining pass after pruning.
Compared to iterative pruning approaches, which often continually fine-tune to compensate for the performance lost due to pruning, \alg{} falls firmly in the domain of single-shot pruning methods.
% Pseudocode for SNACS is presented in Alg.~\ref{alg:SNACS}.

\begin{algorithm}
\SetAlgoLined
 \For{Every pair of layers $(l, l+1), l \in {1,2,\dots,L-1}$}
 {
 Compute $\gamma^{(l+1)}$\;
 \For{$F^{(l+1)}_i, i \in  \{1,2,\dots N^{(l+1)}\} \setminus \textsc{Sensitive\_Filters}\left(\left\{1,2,\dots N^{(l+1)}\right\} \right)$}
%  $S_{F^{(l+1)}_i}$}
 {
 Initialize $S_{F^{(l+1)}_i} =\emptyset$\;
 \For{$F^{(l)}_j, j \in 1,2,\dots N^{(l)}$}
 {
 Compute $\eta(F^{(l+1)}_i, F^{(l)}_j|\overline{ F^{(l)}_j})$\;
%  = $I_{\phi}(F^{(l+1)}_i, F^{(l)}_j\mid \overline{ F^{(l)}_j})$\;
 \If{$\left(\eta{(F^{(l+1)}_i, F^{(l)}_j|\overline{ F^{(l)}_j})} \leq \delta\right.$ \textbf{and} $\left.\sum_i \vert S_{F^{(l+1)}_i}\vert /(N^{(l+1)}N^{(l)}) < \gamma^{(l+1)}\right)$}
 {
 $S_{F^{(l+1)}_i} = S_{F^{(l+1)}_i} \cup \text{index}(F^{(l)}_j)$
 }
 }
 }
 }
\caption{{\alg}  pruning between filters of layers ($l, l+1$)}
\label{alg:SNACS}
\end{algorithm}

\subsection{Adaptive Conditional Mutual Information}
\label{sec:weighted_hash_based_cmi_estimator}

In the following subsection, we introduce Adaptive Mutual Information, a non-linear dependency measure that is based on the $f$-divergence measure~\cite{TC,Csisz1967}, extend it to a conditional formulation and discuss the hash-table-based estimator used to compute ACMI. 

 \hfill \\
\noindent\textbf{Definition}: Let $\mathcal{X}$ and $\mathcal{Y}$ be Euclidean spaces and let $P_{XY}$ be a probability measure on the space $\mathcal{X}\times \mathcal{Y}$. 
Here, $P_X$ and $P_Y$ define the marginal probability measures. 
Similar to \cite{Yurietal2016}, for given function $(x,y)\in \mathcal{X}\times \mathcal{Y} \mapsto \varphi(x,y)\geq 0$, the Adaptive Mutual Information (AMI), denoted by $I_\varphi (X;Y)$, is defined as, 
\begin{equation}\label{def:AMI}
  I_\varphi(X;Y) = \mathop{\mathbb{E}}_{P_X P_Y}\left[ \varphi(X,Y) g\left(\frac{dP_{XY}}{dP_X P_Y}\right)\right], 
\end{equation}
where $\frac{dP_{XY}}{dP_X P_Y}$ is the Radon-Nikodym derivative, and $g: (0,\infty)\mapsto \mathbb{R}$ is a convex function and $g(1)=0$.
Note that when $\frac{dP_{XY}}{dP_X P_Y}\rightarrow 1$ then $I_\varphi\rightarrow 0$. 
The overall bounds on the AMI measure are given by,
\begin{equation}
    0\leq I_\varphi(X,Y)\leq \frac{1}{2}\mathop{\mathbb{E}}_{P_X P_Y}\left[\varphi(X,Y) \left(\frac{dP_{XY}}{dP_X P_Y}+1\right)\right].
\end{equation}
An explanation of how we arrive at these bounds is provided in Appendix A.

%In particular if $g(t) = \frac{(t-1)^2}{2(t+1)}$, we have
%\begin{equation}\label{def:AMI:arlter}
%I_\varphi(X;Y)=\frac{1}{2}\mathop{\mathbb{E}}_{P_X P_Y}\left[\varphi(X,Y) \left(\frac{dP_{XY}}{dP_X P_Y}+1\right)\right]-2 \mathop{\mathbb{E}}_{P_X P_Y}\left[\varphi(X,Y)h\left(\frac{dP_{XY}}{dP_X P_Y}\right)\right], 
%\end{equation}
%where $h(t)=\frac{t}{t+1}$. When $P_{XY}$ and $P_XP_Y$ have no overlapping space then the second term in (\ref{def:AMI:arlter}) becomes zero. Therefore, bounds on $I_\varphi$ is given as,
%\begin{equation}\label{bound.AMI}
%    0\leq I_\varphi(X,Y)\leq \frac{1}{2}\mathop{\mathbb{E}}_{P_X P_Y}\left[\varphi(X,Y) \left(\frac{dP_{XY}}{dP_X P_Y}+1\right)\right].
%\end{equation}
% The AMI measure is extensively discussed in \cite{Yurietal2016}. 
%\vspace{-0.3cm}

\hfill \\
\noindent\textbf{Adaptive Conditional Mutual Information}: 
Let $\mathcal{X}$, $\mathcal{Y}$ and $\mathcal{Z}$ be Euclidean spaces and let $P_{XYZ}$ be a probability measure on the space $\mathcal{X}\times \mathcal{Y}\times \mathcal{Z}$. 
We presume $P_{XY|Z}$, $P_{X|Z}$, and $P_{Y|Z}$ are the joint and marginal conditional probability measures, respectively. 
$P_{Z}$ defines the marginal probability measure on the space $\mathcal{Z}$. 
% the construction of information-theoretic measures in
Following~\cite{Yurietal2016}, the Adaptive Conditional Mutual Information (ACMI), denoted by $I_\varphi(X;Y|Z)$, is defined as,
\begin{equation}\label{def:ACMI}
  I_\varphi(X;Y|Z) = \mathop{\mathbb{E}}_{P_Z P_{X\mid Z} P_{Y\mid Z}}\left[ \varphi(X,Y,Z) g\left( \frac{dP_{XY\mid Z}}{dP_{X\mid Z}P_{Y\mid Z}} \right)\right]. 
\end{equation}
In this paper, we focus on the particular case of $g(t) = \frac{(t-1)^2}{2(t+1)}$ so $I_\varphi \in [0,1]$. 
Note that when $\varphi=1$, the ACMI in (\ref{def:ACMI}) becomes the conditional geometric MI measure proposed in \cite{SalimehEntropy2018}.
Next we propose a hash-based estimator of ACMI to approximate the connectivity score between filters.

 \hfill \\
\noindent\textbf{Hash-based Estimator of ACMI}:
Consider $N$ i.i.d samples $\big\{(X_i,Y_i,Z_i)\big\}_{i=1}^N$ drawn from $P_{XYZ}$,
which is defined on the space $\mathcal{X}\times \mathcal{Y} \times \mathcal{Z}$. 
We define a dependence graph $G(X,Y,Z)$ as a directed multi-partite graph, consisting of three sets of nodes $V$, $U$, and $W$, with cardinalities denoted as $\vert V \vert$, $\vert U\vert$, and $\vert W\vert$, respectively and with the set of all edges $E_G$. 
The variable $W$ here is different from the DNN weight matrix. 
% but in a general fashion 
Following similar arguments to \cite{Mortezaetal2019}, we map each point in the sets $\mathbf{X}=\{X_1,\ldots,X_N\}$, $\mathbf{Y}=\{Y_1,\ldots,Y_N\}$, and $\mathbf{Z}=\{Z_1,\ldots,Z_N\}$
to the nodes in the sets $V$, $U$, and $W$, respectively, using the hash function $H$.

Here, $H(x)=H_2(H_1(x))$, where the vector valued hash function $H_1:\mathbb{R}^d\mapsto \mathbb{Z}^d$ is defined as $H_1(x)=[h_1(x),\ldots,h_1(x_d)]$, for $x=[x_1,\ldots,x_d]$ and $h_1(x_i)=\lfloor\frac{x_i+b}{\epsilon}\rfloor$, for a fixed $\epsilon > 0$, and random variable $b\in[0,\epsilon]$. 
The random hash function $H_2:\mathbb{Z}^d\mapsto\mathcal{F}$ is uniformly distributed on the output $\mathcal{F}=\{1,2,\ldots,F\}$ where for a fixed tunable integer $c_H$, $F=c_H N$. 

After the projection of values on to the dependence graph $G(X,Y,Z)$, we define the following cardinality,
\begin{align}
N_{ijk}= \#\{(X_t,Y_t,Z_t) \;\; s.t.\;\; H(X_t)=i,\\ \notag 
H(Y_t)=j, H(Z_t)=k\},  
\end{align}
which is the number of joint collisions of the nodes $(X_t,Y_t,Z_t)$ at the triple $(v_i,u_j,\omega_k)$.
Let $N_{ik}$, $N_{jk}$, and $N_k$ be the number of collisions at the vertices $(v_i,\omega_k)$, $(u_j,\omega_k)$, and $\omega_k$, respectively. 
By using $N_{ijk}$, $N_{ik}$, $N_{jk}$, and $N_k$, we define the following ratios,
\begin{equation}
    r_{ijk} := \frac{N_{ijk}}{N}, \;\; r_{ik} := \frac{N_{ik}}{N}, \;\;r_{jk} := \frac{N_{jk}}{N}, \;\; r_{k} 
    := \frac{N_k}{N}.
\end{equation}
Finally, using the above ratios we propose the following hash-based estimator of the ACMI measure (\ref{def:ACMI}): 
\begin{equation}\label{Est:ACMI}
    \widehat{I}_{\varphi}(X;Y|Z) = \sum_{e_{ijk}\in E_G}  {{\varphi}}(i,j,k)\; \frac{r_{ik} \; r_{jk}}{r_k}g\left(\frac{r_{ijk}\; r_k}{r_{ik}\; r_{jk}}\right), 
\end{equation}
summed over all edges $e_{ijk}$ of $G(X,Y,Z)$ having non-zero ratios.

\begin{theorem}\label{thm.1}
For given $g(t) = \frac{(t-1)^2}{2(t+1)}$ and under the assumptions: {\bf (A1)} The support sets $\mathcal{X}$, $\mathcal{Y}$, and $\mathcal{Z}$ are bounded. %{\bf (A2)} The supremum $g\left( \frac{dP_{XY\mid Z}}{dP_{X\mid Z}P_{Y\mid Z}} \right)$ exists and is bounded.
{\bf (A2)} The function $\varphi$ is bounded. 
{\bf (A3)} The continuous marginal, joint, and conditional density functions are belong to H\"{o}lder continuous class, \cite{WH1990}. For fixed $d_X$, $d_Y$, and $d_Y$, as $N\rightarrow \infty$ we have
\begin{equation}
    \widehat{I}_{\varphi}(X;Y|Z) \longrightarrow I_\varphi(X;Y|Z), \;\;\; a.s.
\end{equation}
\end{theorem}

The proof of Theorem \ref{thm.1} is available in the Appendices. 

\hfill \\
\noindent\textbf{Implementation}: Overall, $X, Y, Z$ denote sets of activations derived from different filters and we obtain a scalar value (connectivity score) as the outcome of the ACMI estimator in (\ref{Est:ACMI}).
The flexibility in defining function $\varphi$ offers a way to connect the probabilistic framework of MI to existing weight-based pruning approaches.
In Section~\ref{sec:experimental_results}, we explore a variety of options for $\varphi$ and empirically determine that a function defined on the weight matrix helps achieve the highest pruning performance in our experiments.

\subsection{Definition of Upper Pruning Percentage Limit of Layers}
\label{sec:algorithm_defining_upper_pruning_limit_of_a_layer}

\begin{figure}[t]
    \centering
    \includegraphics[width=\columnwidth]{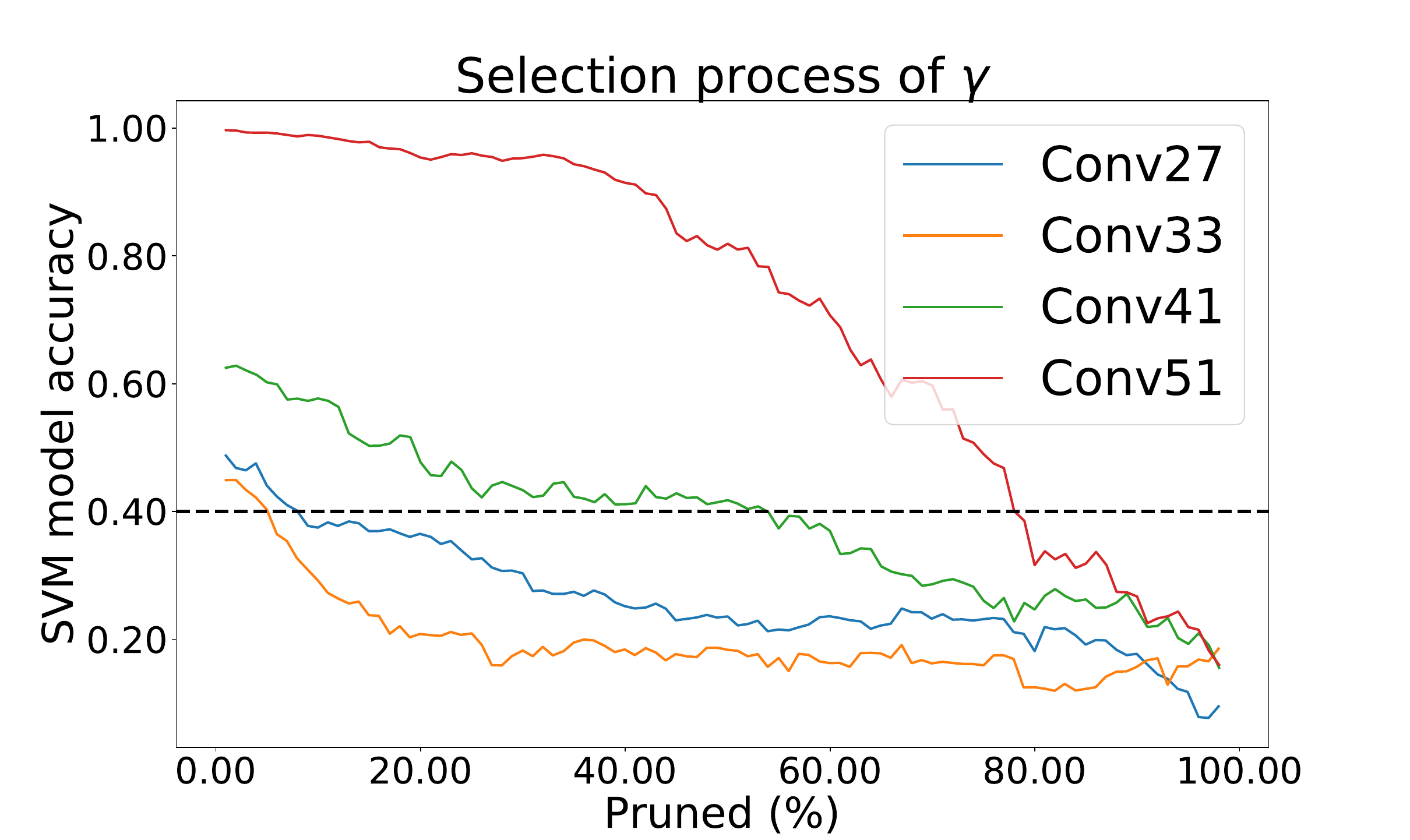}
    \caption{The selection process for upper pruning percentage limits for each layer of the DNN is based on using a fixed threshold (dotted line) over the SVM model's performance such that the weighted sum of Pruned($\%$) allocated to each layer, the x coordinate where the threshold intersects the curve latest, matches the overall sparsity $\tau$.}
    \label{fig:ul}
\end{figure}
% Updated contents
To protect different layers of the DNN from being excessively pruned, we propose a set of operating constraints to automate the joint definition of the upper pruning percentage limits of every layer in the DNN. 
Our approach is based on trends in the degradation of the quality of activations  when a layer is pruned to varying extents.
At each layer, we collect the performances of an SVM model with an RBF kernel ($\alpha^{(l)}_c$), trained on a subset of activations from the un-pruned version of the layer and tested on the same subset from the pruned version of the layer at various compression levels $c$, where $c \in \{1,2,3\dots 99\}$.
Here, the ground-truth labels from the dataset are used to train the SVM model.

Once we have the performance of SVM models across all layers, we cycle through performances between $100-0\%$ to find the optimal threshold value such that the sum of compression levels of all the layers dictated by the selected threshold adds up to our overall target pruning percentage.
Each individual layer's pruning percentage is dictated by the highest compression level where the SVM model's performance exceeds the chosen threshold.
The general trend we observe is higher the compression level, lower is the SVM model's performance.
Thus, picking smaller performance thresholds leads to the selection of higher compression levels in a layer.
We select the highest compression level from a range of possible values to avoid noisy and inconsistent behavior in SVM performances.
Mathematically, we optimize,

% Critical to a well functioning pruning algorithm is the existence of a fail-safe that protects layers from being excessively pruned.
% Often, the fail-safe, in the form of upper pruning percentage limits for each layer in the DNN, is determined through a lot of trial and error.
% To help avoid this, we propose a set of operating constraints to automatically define the upper pruning percentage limits of every layer in the DNN.

% Our approach is based on trends in the degradation of the quality of activations  when a layer is pruned to varying extents.
% First, we pose the desired overall pruning percentage of the DNN, $\tau$, as the weighted sum of upper pruning percentage limits defined on every layer ($\gamma^{(l)}$).
% This is denoted as:
\begin{equation}
    \tau = \sum_{l=1}^{L} \beta^{(l)} \gamma^{(l)},
\end{equation}
where $\beta^{(l)}$ is ratio of number of parameters in layer $l$ to the total number of parameters across the entire DNN and $\tau$ denotes the desired pruning percentage across the entire DNN.
% To find $\gamma^{(l)}$, first, we collect the performance of an SVM model that is trained on activations from layer $l$, prior to being pruned.
% This SVM model is then evaluated on activations collected after the layer is pruned to varying levels.
% Here, the performance of the SVM model for a given layer $l$ at a specific pruning percentage $c$ is denoted by $\alpha^{(l)}_c$.
% After collecting $\alpha$s across every layer in the DNN we determine the pruning threshold for each layer by selecting an $\alpha$ such that the weighted combination of pruning percentages determined by the first point of intersection with each SVM curve, as shown in Fig.~\ref{fig:ul}, match $\tau$.
% The pruning threshold for each layer in the DNN is determined by selecting an $\alpha$ value such that the sum of the weighted combinations of the pruning percentages associated with the selected threshold in each layer is constrained to be $\tau$.
Fig.~\ref{fig:ul} illustrates this process using an example of 4 layers.

It is important to note that the statistics computed from the SVM models across all layers can be executed in parallel, at an average of 36s per SVM model.
This is an important distinction when compared to prior work where optimization involves computing the permutation of pruning percentages across various layers (order of $99^{l}$). 
Across each such permutation, the entire network needs to be retrained/fine-tuned, which can take anywhere from a couple of hours (CIFAR-10) to a week (ILSVRC-2012). This cost is significantly higher when compared to the simple forward pass across the DNN and training time for an RBF-SVM model used in our approach.
Our core contribution in this work is a systematic approach to decide the upper pruning percentage limits across all layers of the DNN.
Previous works often relegate this information to the final chosen values without disclosing how they arrived at them.
% Once all such performances are collected, we create two subsets of layers in the DNN, one in which layers can be maximally compressed, denoted by $M$, and the remaining set of layers, denoted as $\overline{M}$.
% The layers in $M$ are those which contain performances in the top $80\%$, regardless of the pruning percentage at which this performance is achieved. For all layers in $M$, we set $\gamma^{(l)}$ to be the largest pruning percentage that maintains the performance of the layer above random for a given dataset.
% Unlike the individual thresholds used for layers in $M$, the upper pruning percentage limit for the remaining set of layers is computed based on a common threshold over all the evaluated performances in the DNN, $\alpha$, subject to the constraint on $\tau$.
We provide the $\gamma$ values of all layers for each DNN architecture used in the experiments in our supplementary materials.

\subsection{Sensitivity of Filters}
\label{sec:evaluating_sensitivity_of_filters}

A common assumption made during pruning is that all filters in a layer have the same downstream impact and hence can be characterized solely using the magnitude of their weights.
In contrast, probabilistic pruning approaches like MINT~\cite{ganesh2020mint} aim to maintain the flow of information between a pair of layers but they consider all filters to be equally important.
Taking into account each filter's impact on succeeding layers is an effective tool to assess their importance and protect filters that contribute the majority of information from being pruned.

We define a sensitivity criterion, $\lambda(F^{l+1}_{i})$, that can be used to sort filters in their order of importance.
Using this, we curate a subset of filters that are critical and hence need to be protected from pruning while the remaining filters are pruned using the steps in Alg.~\ref{alg:SNACS}.
To evaluate the sensitivity of filters in layer $l+1$, we look at the weight matrix of its downstream layer $l+2$, $W^{(l+2)}$, and assess the contributions from filters in $l+1$ to those in $l+2$.
Here, $W^{(l+2)} \in \mathbb{R}^{N^{(l+2)} \times N^{(l+1)} \times H \times W}$, where $H,W$ are the height and width of the filters in layer $l+2$.
For a given filter, the sum of normalized contributions across all the filters in $l+2$ is its overall sensitivity, $\lambda(F^{(l+1)}_i)$.
It is defined as,
\begin{equation}
    \lambda(F^{l+1}_{i}) = \sum_{f_c=1}^{N^{(l+2)}}\widetilde{W}^{(l+2)}(f_c, i)\big/\mathcal{C}^{(l+2)}(f_c),
    \end{equation}
    \begin{equation}
    \hbox{where}\;\;\mathcal{C}^{(l+2)} (f_c)= \sum_{f_p=1}^{N^{(l+1)}}\widetilde{W}^{(l+2)}(f_c, f_p).
\end{equation}
Here, $\mathcal{C}^{(l+2)}$ is the normalization constant used to relate the weights of filters from $l+1$ contributing to the same filter in $l+2$ and $\widetilde{W}^{(l+2)}$ is the weight matrix of $l+2$ averaged over the height and width.

Once we obtain the order of sensitivity values for filters in a given layer, we define a threshold of highly sensitive filters that remain un-pruned, after empirically comparing the improvement in performance at similar pruning levels with and without protecting sensitive filters.
This is critical to ensure that only sensitive filters, which contribute a majority of the information downstream, remain untouched.
This in turn helps improve the overall compression performance since less sensitive filters can be pruned more without compromising the quality of information flowing between layers too much.
% Finally, we send the indices of sensitive filters that need to be protected to Alg.~\ref{alg:SNACS}.
After empirically comparing the degradation in performance of the SVM model used to define the upper pruning percentage limits for layers, between the case when all the filters are pruned and when we protect a variable percentage of sensitive filters, we determine the set of highly sensitive filters to protect and return their indices to Alg.~\ref{alg:SNACS}.

% \textcolor{red}{Can probably be moved to appendix or these points can be put elsewhere?}
% While the overall construction of our pruning algorithm is similar to \cite{ganesh2020mint}, we highlight some key differences between the two approaches.
% \begin{itemize}
% \item We derive and use a plugin hash-based estimator instead of an MST-based estimator. This offers a higher degree of accuracy and significant improvement in run-time.
% \item We use a custom algorithm to define the upper pruning limit for every layer in a neural network while they manually define it.
% \item Using our notion of sensitivity, we show that protecting a fraction of sensitive filters in a layer can further be used to boost pruning.
% \end{itemize}

\section{Experimental Results}
\label{sec:experimental_results}
We divide our results into three subsections, formatted as an ablative study.
Section~\ref{sec:estimator_results} focuses on the evaluation of run-time and choice of $\varphi$, to highlight the impact of using our ACMI estimator in place of the MST-based estimator used in \cite{ganesh2020mint}. 
Here, the upper pruning limits are manually defined, with the help of artificial limits placed on the SVM model accuracy, to mimic prior work.
In Section~\ref{sec:main_results}, we detail the results of applying \alg{} (ACMI + Automated upper pruning percentage limits) across three Dataset-DNN combinations.
Within this section we focus on drawing strong comparisons against single-shot pruning approaches while also highlighting how competitive \alg{} is amongst approaches that use a modified objective function or iterative pruning.
Finally, in Section~\ref{sec:sentivity_results} we discuss the impact of adding our sensitivity measure as a way to prioritize and fully protect important filters from being pruned.

\hfill \\
\noindent\textbf{Dataset-DNN}:
We use three standard Dataset-DNN combinations to evaluate and compare our approach to standard baselines.
They are, CIFAR10~\cite{krizhevsky2009learning}-VGG16~\cite{simonyan2014very}, CIFAR10-ResNet56~\cite{He_2016} and ILSVRC2012~\cite{ILSVRC15}-ResNet50.
A detailed breakdown of each dataset and the experimental setup used in each experiment is included in the supplementary materials.

\hfill \\
\noindent\textbf{Metric}:
We use the following metrics to compare performances,
\begin{itemize}
    \item Pruning~($\%$): The percentage of parameters removed when compared to the total number of parameters in the un-pruned DNN (Conv and FC only),
    \item Test Accuracy~($\%$): The accuracy on the testing set, after re-training for pruned networks,
    \item Memory~(Mb): The amount of memory consumed to store the weight matrices in ``CSR'' format.
    \item FLOPs Reduced~($\%$): The percentage of FLOPs reduced when compared to the un-pruned DNN.
\end{itemize}
Apart from the above metrics, we also use \textit{run-time} to compare speed of estimators.
A high quality method must have high compression performance while maintaining a test accuracy relatively close to the baseline.

\subsection{Evaluation of Estimator}
\label{sec:estimator_results}

\begin{figure}[t!]
    \centering
    \subfloat[][Run-time comparison across MST and ACMI measures]{\includegraphics[width=\columnwidth]{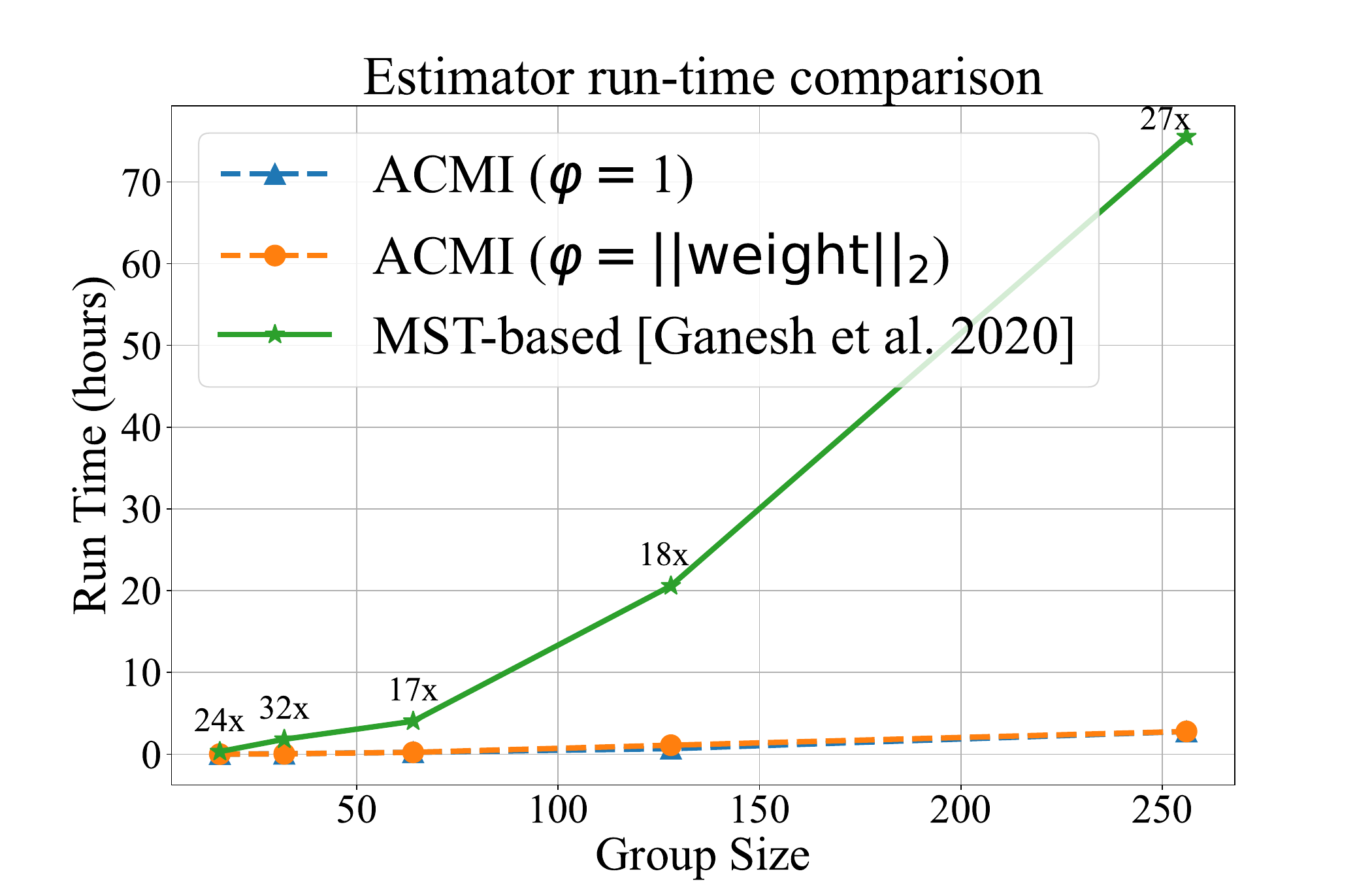}\label{fig:mint_snacs_runtime}}
    
    \subfloat[][Run-time comparison across various $\varphi$]{\includegraphics[width=\columnwidth]{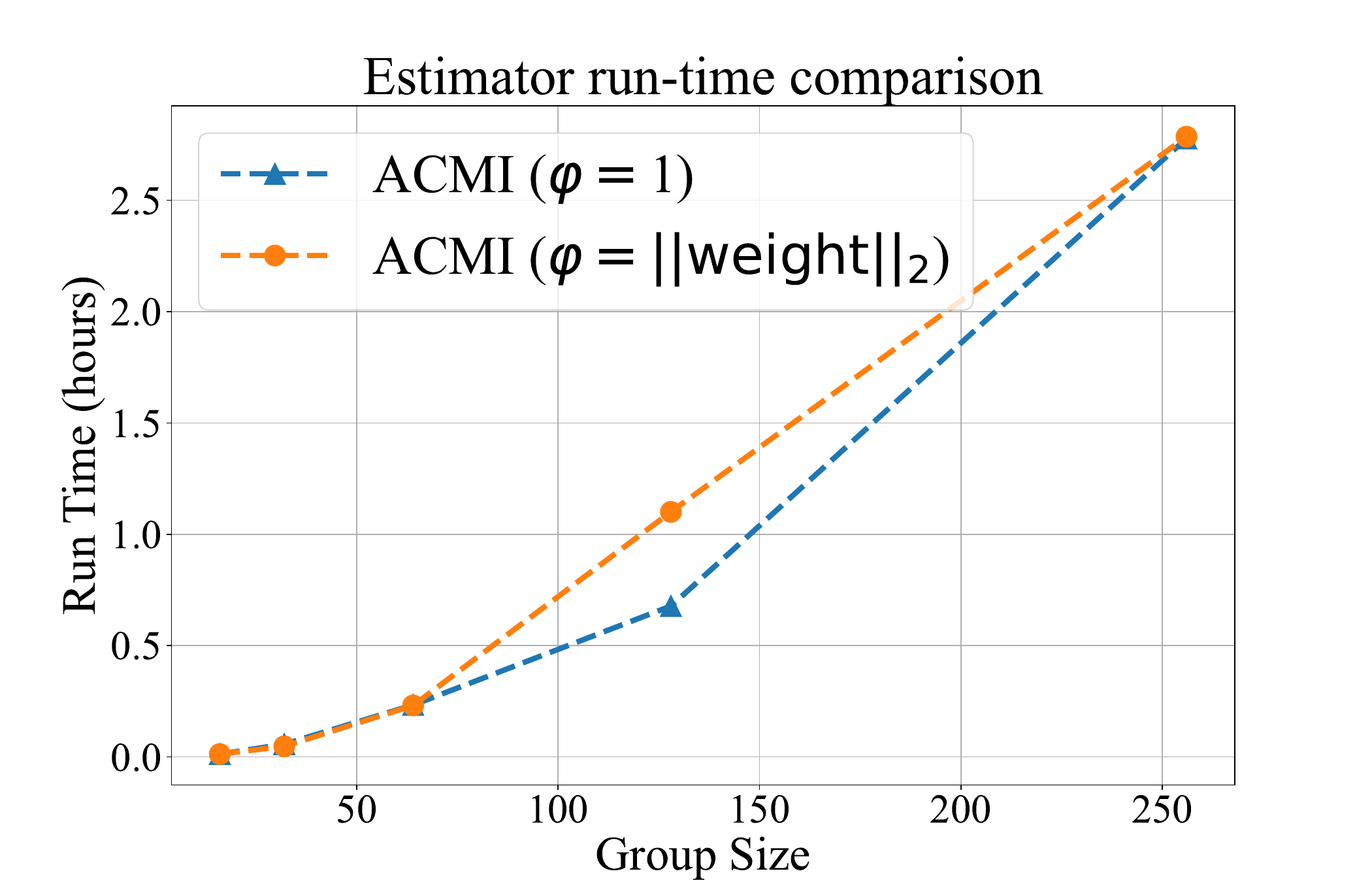}\label{fig:snacs_runtime}}
    \caption{(\ref{fig:mint_snacs_runtime}) When comparing run-times between the MST-based estimator used in MINT~\cite{ganesh2020mint} and our hash-based ACMI estimator, our estimator provides up to $27\times$ speedup in run-time. (\ref{fig:snacs_runtime}) Across different selections of the scaling function in our estimator, the run-times scale similarly as the number of groups increase.}
    \label{fig:runtime}
\end{figure}

\noindent\textbf{Run-time Comparison}:
We provide a comparison between the run-time taken to compute the dependency scores across convolution layer 9 in VGG16 using our proposed ACMI estimator and the MST-based estimator used in MINT~\cite{ganesh2020mint}.
For this experiment, we use three distinct estimators, the MST-based estimator from MINT, our ACMI estimator with $\varphi=1$ and $\varphi=\norm{\textrm{weight}}_2$.
Here, $\textrm{weight}$ values are re-scaled between $[0,1]$.
To provide a fair comparison, we adopt the grouping concept introduced in MINT.
From Fig.~\ref{fig:runtime} we make two important observations, 1) run-time increases with an increase in group-size across both estimators, and 2) relative to the run-time from the MST-based estimator, our estimator is faster by at least $17\times$.
Thus, we show that our estimator significantly reduces the overall run-time required to compute conditional MI across a DNN. 
Further, the run-time for one of the largest computational bottlenecks is massively reduced irrespective of the scaling function used in ACMI.

\hfill \\
\noindent{\bf{Selection of $\varphi$}:}
There are number of potential functions we can associate with $\varphi$. 
In Table~\ref{table:phi_functions}, we illustrate a variety of functions and their performance, w.r.t. the Pruning ($\%$) while maintaining an accuracy $\geq 93.43\%$ in the VGG16-CIFAR10 setup.
Between Section~\ref{sec:estimator_results} and MINT~\cite{ganesh2020mint} the main differences are the inclusion of ACMI and the manual definition of upper pruning percentage limits using artificially capped SVM model accuracies (0.8).
% The entire list of functions we tested were derived from $\{\textrm{weights}, \textrm{activations}\} \times \{1, \textrm{exponent}, \textrm{quadratic}\}$ and are provided in the supplementary materials.
% We provide the entire list of functions explored in the Supplementary Materials.
From Table~\ref{table:phi_functions}, we observe that most variants of $\varphi$ outperform MINT, including $\varphi=1$.
Furthermore, we find that $\varphi = \exp(\frac{-\textrm{weights}^2}{2})$ performs the best when compared to all the options for $\varphi$ we explore. 
Thus, we set this as the default $\varphi$ throughout all  further experiments.
% \textbf{Note}: The inclusion of ACMI alongside the manual definition of upper pruning percentage limits using artificially capped SVM model accuracies (0.8) form the main pipeline distinctions between MINT~\cite{ganesh2020mint} and \alg{} in this experiment. 

\begin{table}[t]
\caption{We compare the maximum compression performance of a variety of $\varphi$ functions when maintaining a test accuracy $\geq 93.43\%$. $\varphi = \exp(\frac{-\textrm{weights}^2}{2})$ performs the best and we use this in all further experiments}
\label{table:phi_functions}
\centering
\begin{tabular}{@{}lc@{}}
\toprule
$\varphi$ function & Pruned~($\%$)\\
\toprule
$\textrm{constant} = 1$                                         & 84.02 \\
$\norm{\textrm{weights}}_2$                                     & 84.12 \\
$\textrm{weights}^2$                                            & 84.17 \\
$\exp(\frac{-\textrm{weights}^2}{2})$                           & \textbf{84.46} \\
$\norm{\textrm{act}}_2$                                         & 76.13 \\   
% $\norm{\textrm{act}}_2^2$                                       & --.-- & --.-- \\   
% $\exp(-\frac{\norm{\textrm{act}}_2^2}{2})$                      & --.-- & --.-- \\
$\norm{\textrm{weights}}_2 \norm{\textrm{act}}_2$               & 82.59 \\   
% $\textrm{weights}^2 \norm{\textrm{act}}_2^2$                    & --.-- & --.-- \\   
$\exp(-\frac{\textrm{weights}^2\norm{\textrm{act}}_2^2}{2})$    & 76.99 \\
\bottomrule
\end{tabular}
\end{table}

\subsection{Large-scale Comparison}
\label{sec:main_results}

\begin{table*}[t]
\caption{Using a \textbf{single} train-prune-retrain cycle, \alg{} is among the top performers across all the Dataset-DNN combinations. Baselines are ordered according to increasing Pruning ($\%$)}
\centering
\begin{tabular}{@{}llccc@{}}
\toprule
                              & Method                   & Pruning~($\%$)& Test Accuracy~($\%$) & FLOPs Reduced~($\%$)\\\midrule
\multirow{7}{*}{\shortstack[l]{CIFAR-10 \\VGG16}} & Baseline & N.A.         & 93.98 & N.A. \\ 
& $l_1$-norm~\cite{DBLP:conf/iclr/0022KDSG17}      & 64.00       & 93.40 & 34.18 \\
& Variational Pruning~\cite{zhao2019variational}   & 73.34       & 93.18 & 39.29 \\
& SSS~\cite{huang2018data}                         & 73.80       & 93.02 & 41.60 \\
% & GAL~\cite{lin2019towards}                        & 82.20       & 93.42  & N.A. \\
% & Try-and-Learn~\cite{huang2018learning}           & 82.71       & 91.67  & N.A. \\
& MINT~\cite{ganesh2020mint}                       & 83.46       & 93.43 & N.A. \\
& Network Slimming~\cite{liu2017learning}          & 88.52       & 93.80 & 50.94 \\
& X-Nets~\cite{prabhu2018deep}                     & 92.33       & 93.00 & N.A. \\
% & Cascaded Pruning~\cite{miles2020cascaded}        & 93.52       & 92.97  & N.A. \\
% & DCP~\cite{zhuang2018discrimination}              & 93.58       & 94.57  & N.A. \\
& Bayesian Compression~\cite{louizos2017bayesian}  & 94.50       & 91.00 & N.A. \\
% & RBP~\cite{zhou2019accelerate}                    & 97.46       & 91.00  & N.A. \\\cline{2-5}
& \textbf{SNACS}                                   & 96.16       & 91.06 & 67.85 \\
\midrule
\multirow{5}{*}{\shortstack[l]{ CIFAR-10\\ResNet56 }}  & Baseline & N.A.     & 92.55 & N.A. \\ 
& $l_1$-norm~\cite{DBLP:conf/iclr/0022KDSG17}       & 13.70   & 93.06 & 27.28 \\
& Variational Pruning~\cite{zhao2019variational}    & 20.49   & 92.26 & 20.17 \\
& NISP~\cite{yu2018nisp}                            & 42.60   & 93.01 & 43.61 \\
& FSDP~\cite{gkalelis2020fractional}                & 50.00   & 92.64 & N.A. \\
& MINT~\cite{ganesh2020mint}                        & 57.01   & 93.02 & N.A. \\
% & DHP~\cite{li2020dhp}                              & 58.90   & 92.94 & N.A.\\
% & OED~\cite{wang2019pruning}                        & 60.00   & 92.29 & N.A.\\
% & GAL~\cite{lin2019towards}                         & 65.90   & 91.58 & N.A.\\
% & Paying Attention~\cite{le2020paying}              & 67.05   & 93.34 & N.A.\\
% & DCP~\cite{zhuang2018discrimination}               & 70.32   & 93.81 & N.A.\\
% & Learning Compression~\cite{carreira2018learning}* & 85.01   & 93.08 & N.A.\\
% & Learning Compression~\cite{carreira2018learning}* & 90.01   & 93.33 & N.A.\\
% & Learning Compression~\cite{carreira2018learning}* & 95.00   & 92.49 & N.A.\\
% & Learning Compression~\cite{carreira2018learning}* & 97.00   & 91.79 & N.A.\\\cline{2-5}
& \textbf{SNACS}                                    & 68.59   & 93.38 & 37.61 \\
\midrule
\multirow{5}{*}{\shortstack[l]{ILSVRC2012\\ ResNet50 }}  & Baseline & N.A.     & 76.13  & N.A. \\ 
& SSS~\cite{huang2018data}                  & 38.82  & 71.82 & 43.04 \\
& NISP~\cite{yu2018nisp}                    & 43.82  & 71.99 & 44.01 \\
% & OED~\cite{wang2019pruning}                & 47.37  & 71.22  & N.A.\\
& MINT~\cite{ganesh2020mint}                & 49.62  & 71.05 & N.A. \\
& X-Nets~\cite{prabhu2018deep}              & 50.00  & 72.85 & 50.00 \\
% & DCP~\cite{zhuang2018discrimination}       & 51.45   & 74.95  & N.A.\\
% & \textbf{SNACS}                            & 50.41  & 71.64 & 36.47 \\
& \textbf{SNACS}                            & 55.10  & 74.65 & 41.73 \\
& \textbf{SNACS}                            & 59.61  & 73.60 & 46.63 \\
& \textbf{SNACS}                            & 64.26  & 72.90 & 51.65 \\
& \textbf{SNACS}                            & 68.80  & 72.36 & 56.79 \\
% & GAL~\cite{lin2019towards}                 & 59.96   & 69.31  & N.A.\\
% & ThiNet~\cite{luo2017thinet}               & 66.11   & 68.42  & N.A.\\
% & Cascaded Pruning~\cite{miles2020cascaded} & 72.07   & 72.53  & N.A.\\
% & AutoML~\cite{he2018amc}                   & 80.00   & 76.11  & N.A.\\
% & N2NSkip~\cite{subramaniam2019n2nskip}     & 80.00   & 72.09  & N.A.\\
                            \bottomrule
\end{tabular}
\label{tab:results_main}
\end{table*}

\begin{figure*}[ht!]
    \centering
    \subfloat[][CIFAR10-VGG16]{\label{fig:cifar10_vgg_comp_perf}\includegraphics[width=0.95\columnwidth]{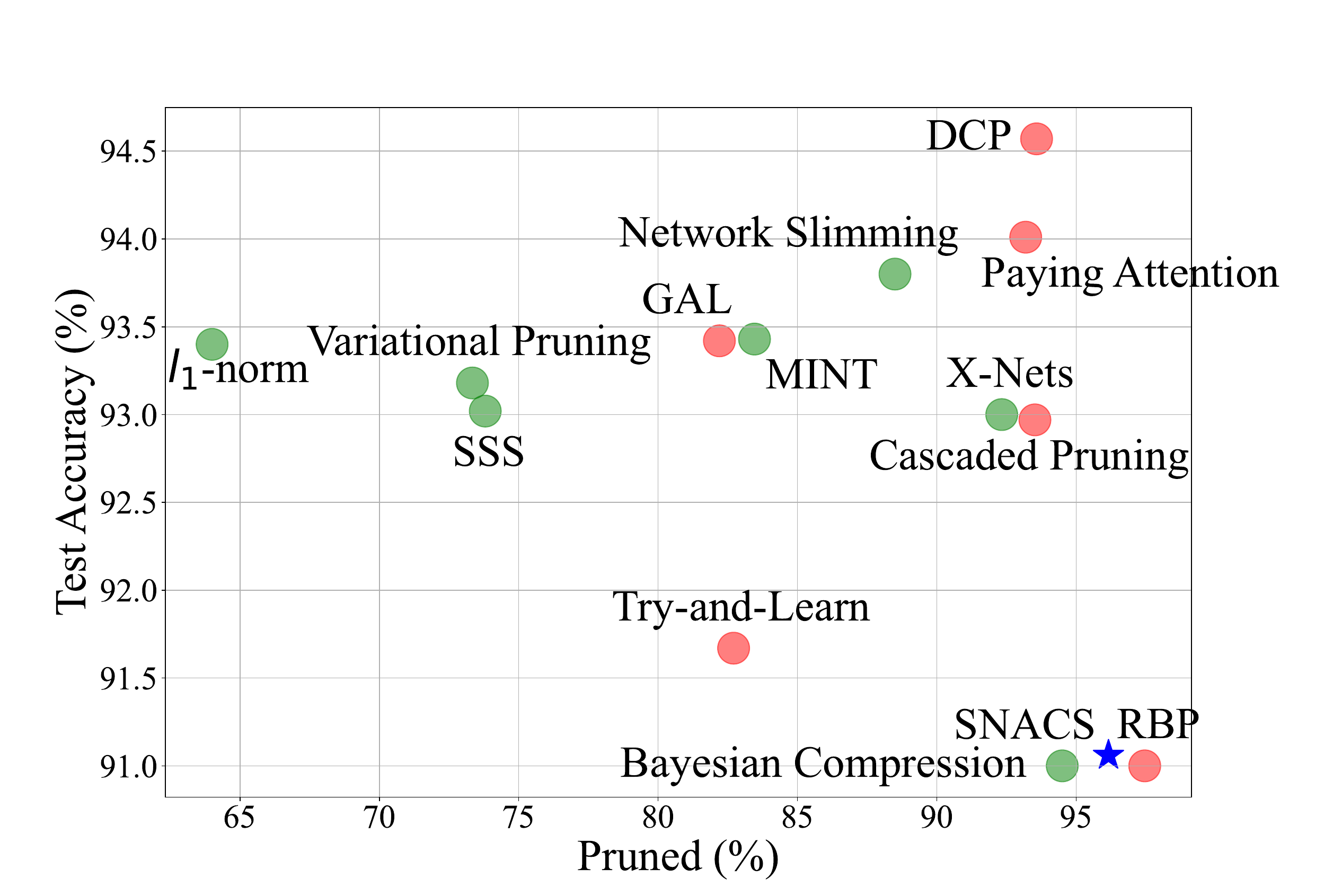}}
    \subfloat[][CIFAR10-ResNet56]{\label{fig:cifar10_r56_comp_perf}\includegraphics[width=0.95\columnwidth]{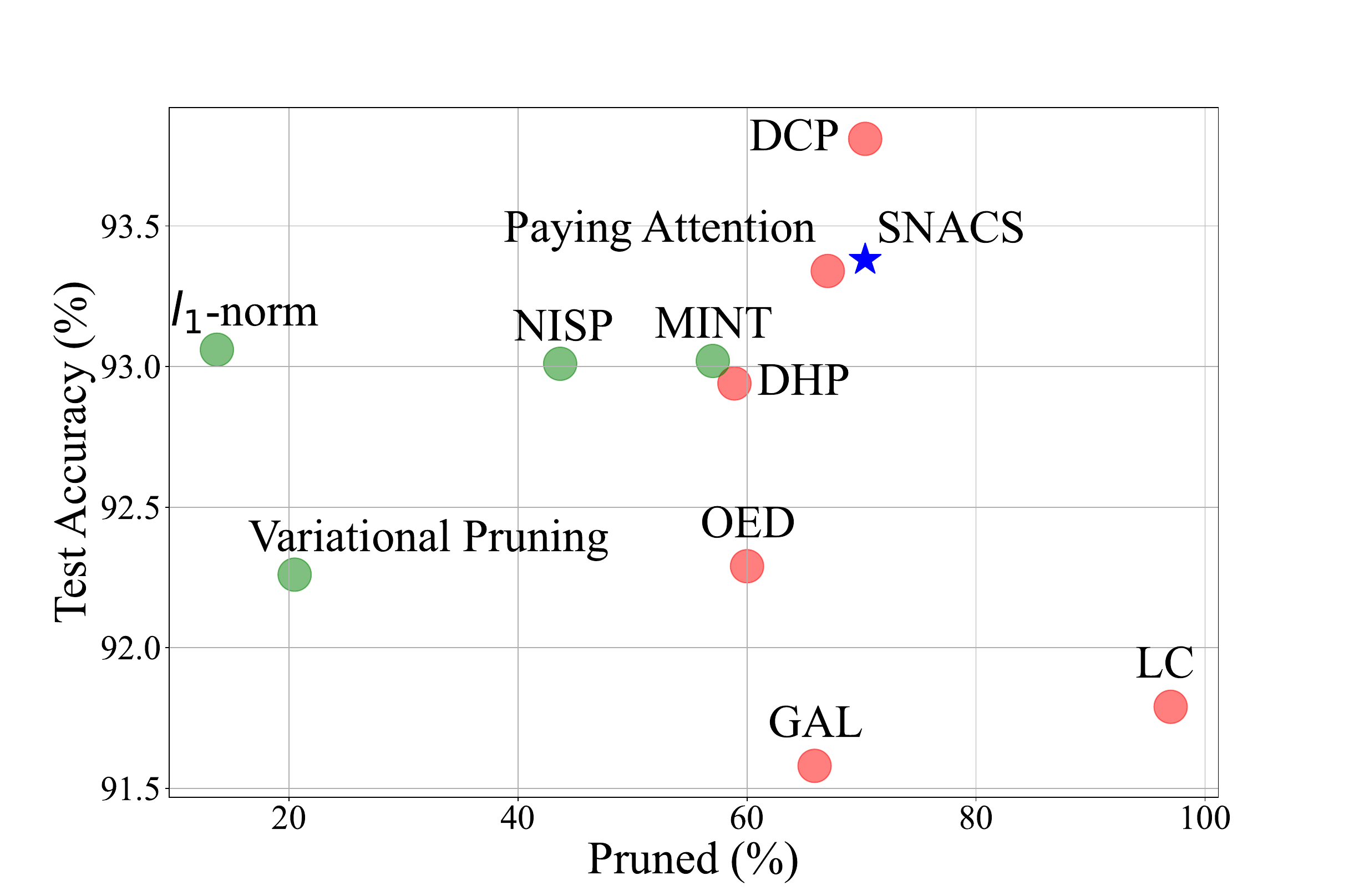}} \\
    \subfloat[][ILSVRC2012-ResNet50]{\label{fig:ilsvrc2012_r50_comp_perf}\includegraphics[width=0.9\columnwidth]{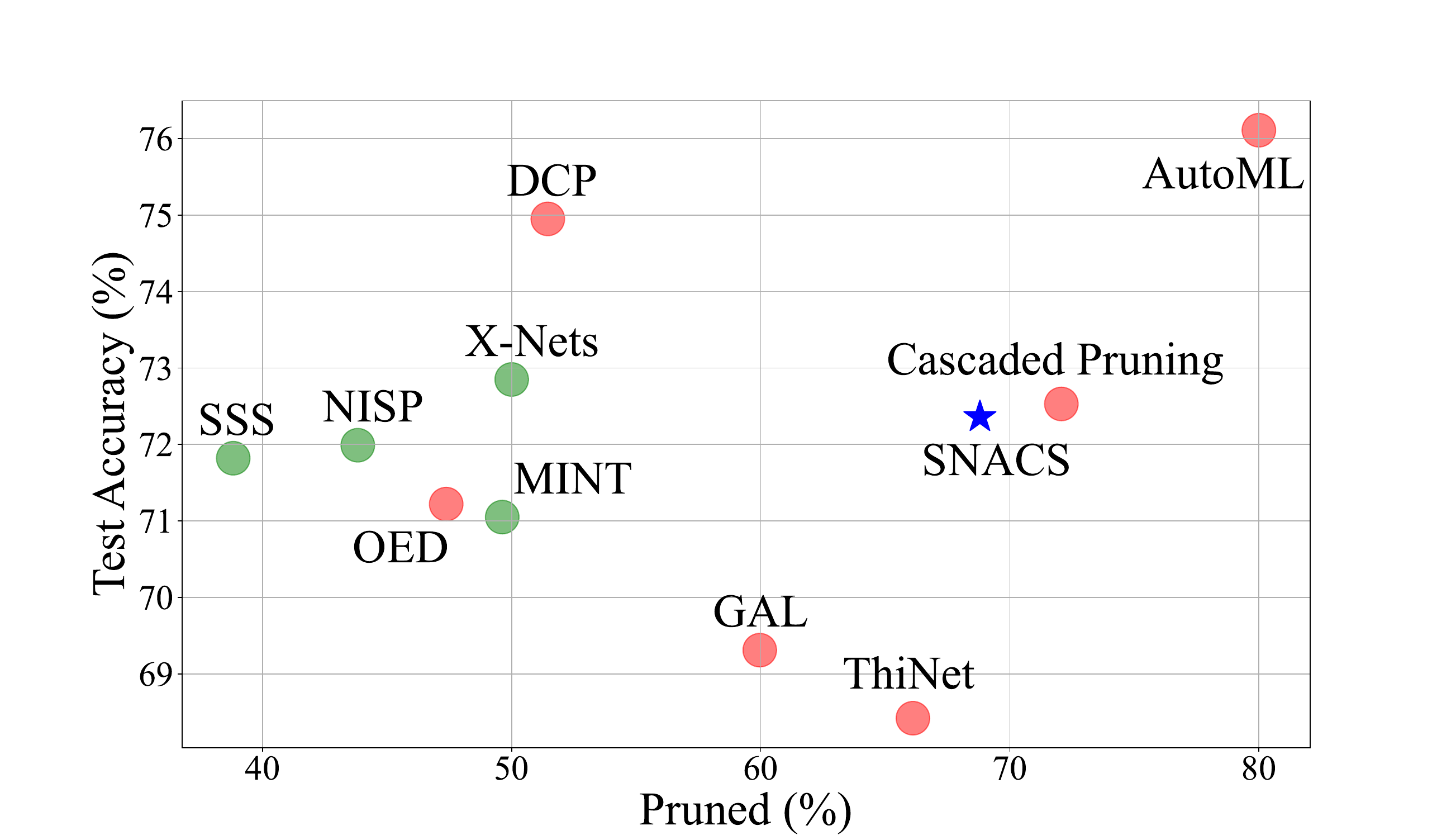}}
    % \subfloat[][]{\label{fig:cifar10_res_comp_per_layer}\includegraphics[width=2\columnwidth]{images/CIFAR10_RESNET56_comp_per_layer.pdf}}\\[-5ex]
    % \subfloat[][]{\label{fig:ilsvrc2012_comp_per_layer}\includegraphics[width=2\columnwidth]{images/ILSVRC2012_comp_per_layer.pdf}}
    \caption{Comparison of single-shot (green) vs. non single-shot (red) pruning approaches across our benchmarks. \alg{}, despite being a single-shot approach, is highly competitive with the best performing iterative methods.}
    \label{fig:com_large_scale}
\end{figure*}
\begin{figure*}[t!]
    \centering
    \includegraphics[width=2\columnwidth]{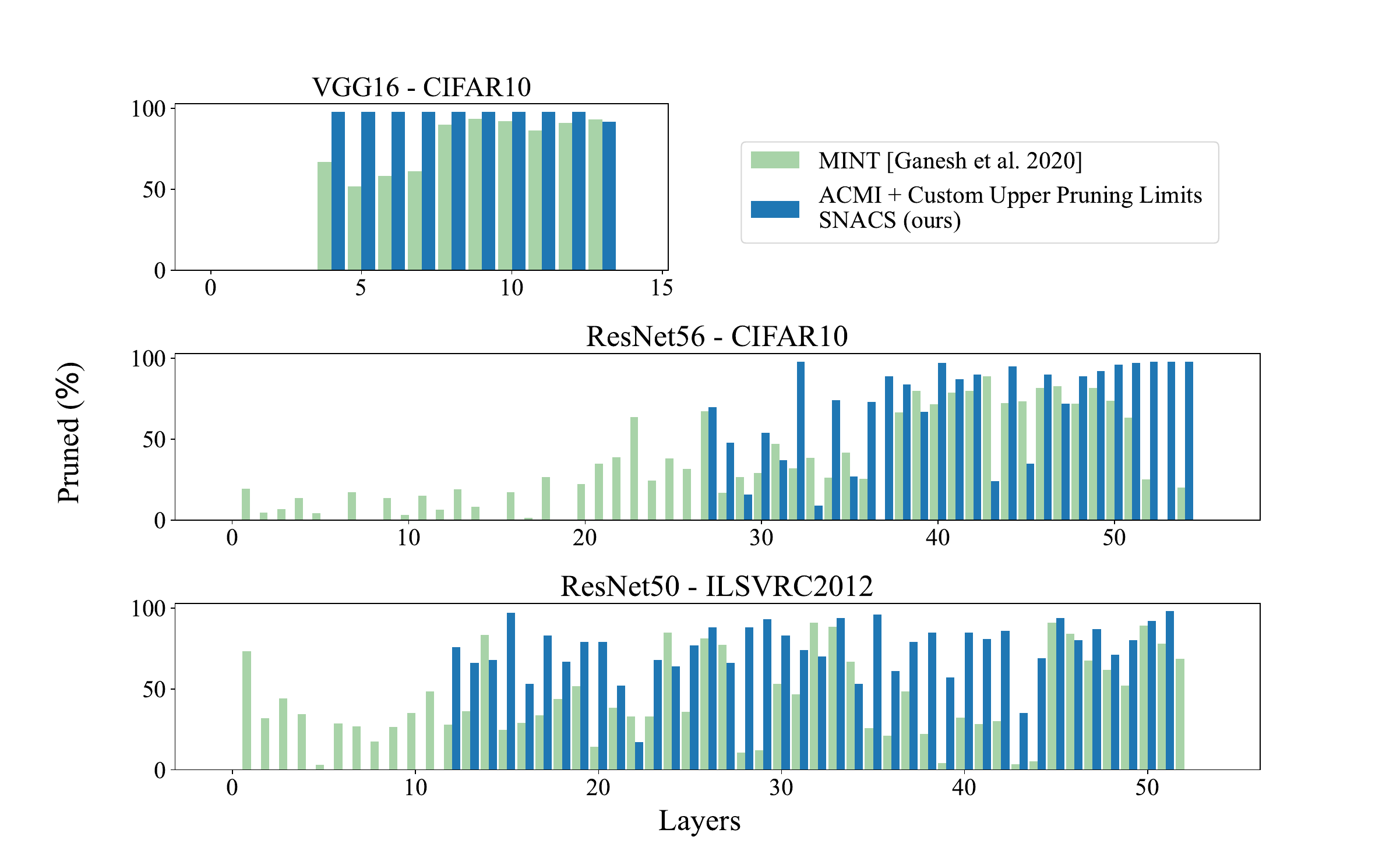}
    % \subfloat[][]{\label{fig:cifar10_vgg_comp_per_layer}\includegraphics[width=2\columnwidth]{images/CIFAR10_VGG16_comp_per_layer.pdf}}\\[-5ex]
    % \subfloat[][]{\label{fig:cifar10_res_comp_per_layer}\includegraphics[width=2\columnwidth]{images/CIFAR10_RESNET56_comp_per_layer.pdf}}\\[-5ex]
    % \subfloat[][]{\label{fig:ilsvrc2012_comp_per_layer}\includegraphics[width=2\columnwidth]{images/ILSVRC2012_comp_per_layer.pdf}}
    \caption{On observing the compression performance per layer in the ILSVRC2012-ResNet50 experiment, \alg{} is able to achieve high Pruning~($\%$) while focusing only on the middle and latter layers and avoiding the early layers. Interestingly, the pattern of pruning in MINT and \alg{} is extremely different.} 
    \label{fig:comp_per_layer_figures}
\end{figure*}
\begin{table*}[ht!]
\caption{By saving a small percentage of sensitive filters, we can further improve the overall Pruning~($\%$) while maintaining high Test Accuracy}
\centering
\begin{tabular}{@{}llcc@{}}
\toprule
                              & Method                   & Pruning~($\%$)& Test Accuracy~($\%$)\\\midrule
\multirow{3}{*}{\shortstack[l]{ResNet56 \\CIFAR-10}} & Baseline & N.A.         & 92.55\\ 
                              & \alg{} (ours)  & 68.59     & 93.38\\
                              & \textbf{\alg{} + sensitivity} (\textbf{ours})  & \textbf{68.96}     & \textbf{93.41}\\
                            \bottomrule
\end{tabular}
\label{tab:sens_results}
\end{table*}
When compared to existing single-shot pruning methods, from Table~\ref{tab:results_main} we observe that \alg{} outperforms all of them by a significant margin to establish new SOTA performances.
Our consistently high results establish our hybrid pruning framework as one of the top performing single-shot algorithms.
A combination of improved estimates from the hash-based ACMI estimator (Table~\ref{table:phi_functions}) and the joint definition of upper pruning percentage limits for each layer in the DNN are the main contributors to our high performance.

Fig.~\ref{fig:com_large_scale} helps put \alg's performance in perspective of pruning approaches that use either sparsity inducing objective functions or iterative re-training setups.
In general, we expect a decrease in performance with an increase in the number of parameters pruned.
Often, iterative approaches achieve the highest compression while suffering minimal drop in testing accuracy, with methods that use joint optimization sprinkled across the entire range of Pruning ($\%$) values.
Single-shot methods are often the weakest performers given that they get the fewest attempts to account for the loss in accuracy after pruning. 
However, across each dataset-DNN combination, our algorithm is highly competitive with the best pruning approaches regardless of variations in optimizers, iterative pruning pipelines, modified objective functions or layer-by-layer fine-tuning.
\alg{} remains competitive at large pruning levels despite using  a \textbf{single} prune-retrain step.

An important distinction between our pruning approach and other single-shot methods we compare against is that we avoid pruning early layers to a large extent, as shown in Fig.~\ref{fig:comp_per_layer_figures}.
Given that a large portion of FLOPs are concentrated in the early portion of the network, the percentage of FLOPs reduced by our \alg{} is slightly lower when compared to methods like X-Nets, which preemptively prunes the network before training, or SSS, which optimizes a different objective function altogether.
% In Section~\ref{sec:estimator_results} we observed the improvement offered by the base ACMI estimates as well as different $\varphi$ functions. 
% Further, we observe a distinct difference in the pattern of $\gamma$'s when compared to those in MINT.  
% Apart from the overall performance achieved by \alg{}, we observe distinct patterns in the definition of  $\gamma^{(l)}$ across all our evaluated benchmarks.
% Here, we take a closer look at the patterns of $\gamma^{(l)}$ obtained from our setup and compare it to MINT to gain further insight into how \alg{} functions.
% From Fig.~\ref{fig:comp_per_layer_figures} we observe that our approach yields a \textbf{distinct and consistent pattern of upper pruning percentage limits that is applicable over all Dataset-DNN combinations} when compared to MINT.
% Comparing the patterns between MINT and \alg{} 
Interestingly, on closer inspection of Fig.~\ref{fig:comp_per_layer_figures}, we observe minimal correlation between the patterns of high and low $\gamma$ values achieved in MINT and our work.
While MINT showcases minimal pruning in the early and middle set of layers, \alg{} focuses on the middle and final set of layers, avoiding the early layers. 
We believe this variation stems from the fact that $\gamma$ values in MINT were co-opted from prior works where the focus on individual layers while in \alg{} the joint definition of $\gamma$s helps capture trends across multiple layers while trying to optimize the performance-sparsity tradeoff.

We observe that when using \alg{} DNNs are more forgiving when pruning layers closer to the output than input since the retraining phase allows them to overcome the loss of abstract concepts learned in later layers but not fundamental structures, when compressing the earlier layers of the network.
Our observations are matched by the discriminant scores in \cite{gkalelis2020fractional} and the median oracle ranking statistics per layer from \cite{molchanov2019pruning}. 
% An important outcome of our general approach to defining upper pruning limits across DNNs is that we make no special provisions when handling different DNN architectures. 
% Often, in prior works, there are certain layers in the middle of the DNN that are explicitly protected from pruning after manually analyzing the impact caused by pruning them~\cite{DBLP:conf/iclr/0022KDSG17} or extra losses are added when processing networks of varying sizes~\cite{zhuang2018discrimination}.
% We observe that \alg{} emphasizes pruning larger portions of the network closer to the output nodes.
However, these observations are in direct contrast to previous works which identify that portions of the network closer to the input are often pruned first~\cite{huang2018data,lin2019towards}.
We hypothesize that their outcomes stem from the modification of the objective function and subsequent training of baseline networks whereas our approach and those in \cite{gkalelis2020fractional,molchanov2019pruning} focus on removing filters based on a pre-defined criterion without the modification of the loss function.
% Further work highlighting the distinctions between methods that modify their objective function and those that prune weights directly is beyond the scope of this work.

\subsection{Sensitivity-based Pruning}
\label{sec:sentivity_results}

\begin{figure*}[ht!]
    \centering
    \subfloat[][Convolution 35]{\label{fig:sens_mask1}\includegraphics[width=\columnwidth]{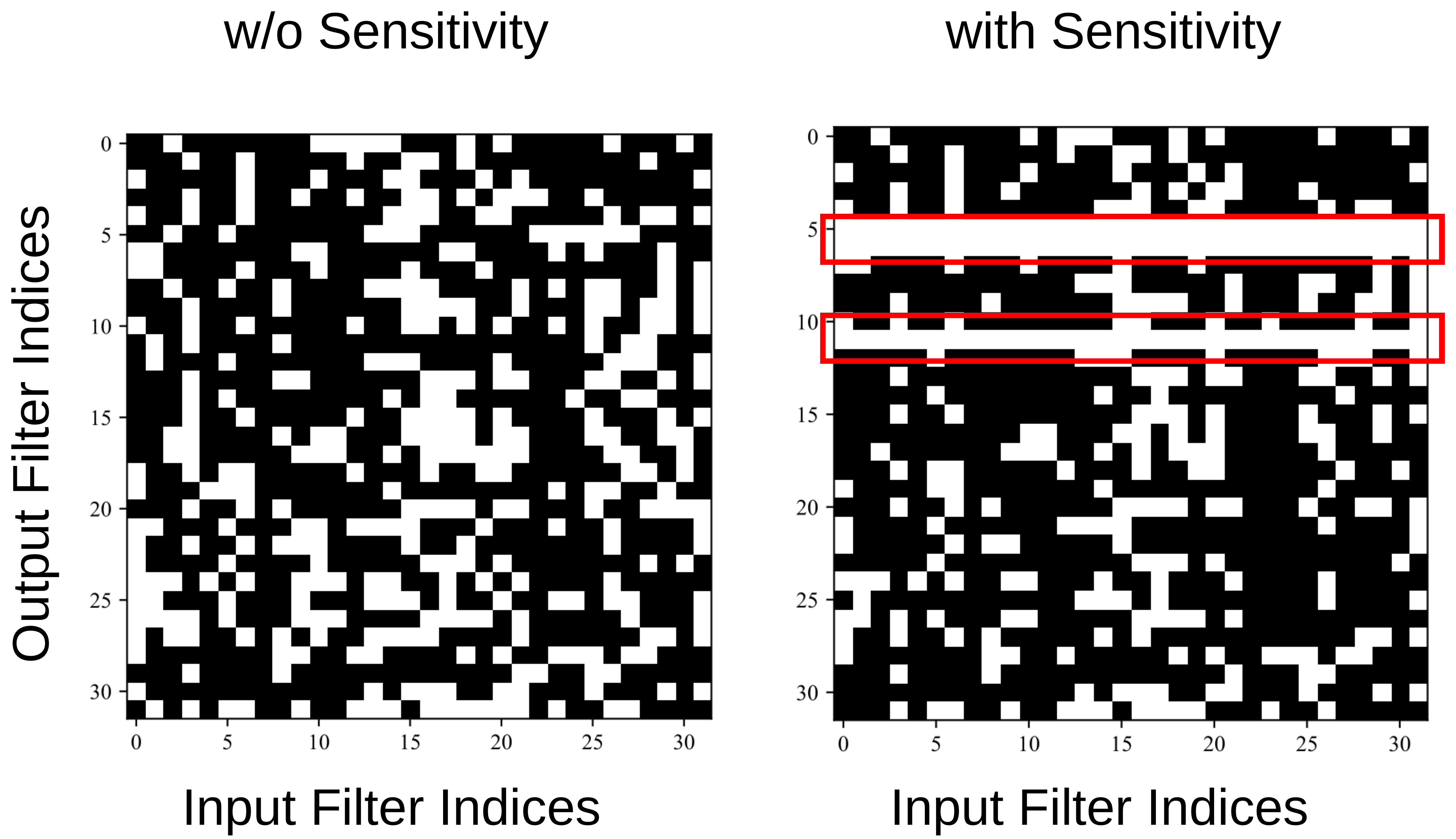}\label{fig:conv4_sens_mask}} 
    \subfloat[][Convolution 46]{\label{fig:sens_mask2}\includegraphics[width=\columnwidth]{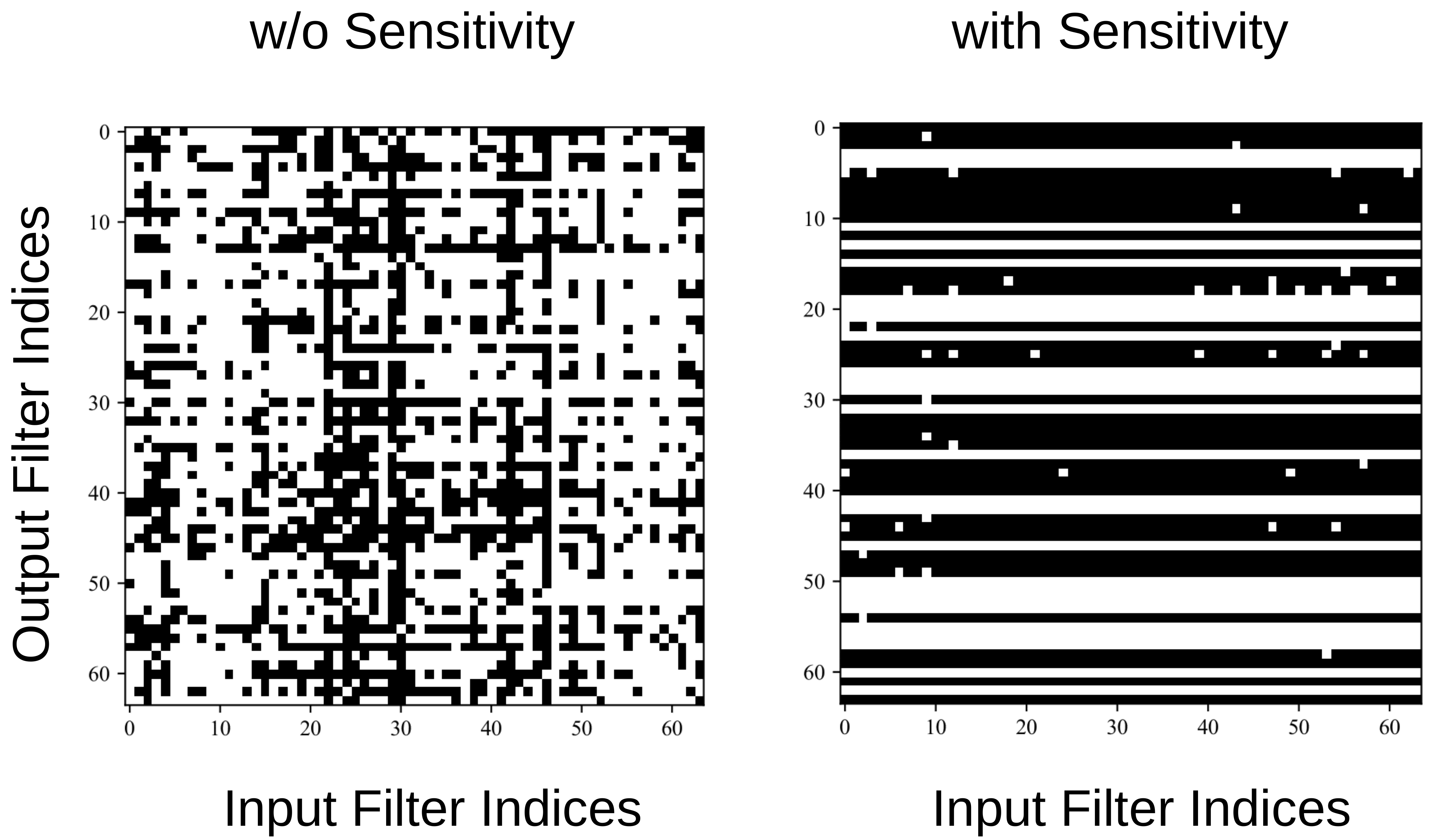}\label{fig:conv9_sens_mask}}
    \caption{Illustrations of filters retained (white) and pruned (black) w/o and with sensitivity based pruning. When protecting important filters from pruning, all its associate connections are maintained (red highlight). An interesting impact of sensitivity is that the connections pruned can be completely modified compared to its counterpart w/o pruning. This is illustrated by the pruning mask of convolution 46.}
    % We attach the plots for the remaining datasets in the supplementary material.}
    \label{fig:combined_figures}
\end{figure*}
\begin{table}[t!]
\caption{Deviating the $\%$ of filters saved from our optimal constraints forces lower sparsity levels with bad testing performance. Optimal values are highlighted in bold}
\centering
\begin{tabular}{@{}cccc@{}}
\toprule
Layer & $\%$ Saved  & Sparsity~($\%$)& Test Accuracy~($\%$)\\\midrule
\multirow{7}{*}{Layer 28} & 30 & 15.03 & 92.83 \\
& 34 & 15.03 & 93.05 \\
& 38 & 14.35 & 92.88 \\
& \textbf{45} & \textbf{55.07} & \textbf{93.41} \\
& 50 & 48.92 & 92.71 \\
& 54 & 45.89 & 93.24 \\
& 60 & 39.74 & 93.10 \\
\midrule
\multirow{7}{*}{Layer 44} & 25 & 26.97 & 92.97 \\
& 30 & 54.83 & 93.13 \\
& 35 & 48.55 & 93.28 \\
& \textbf{40} & \textbf{58.17} & \textbf{93.41} \\
& 45 & 53.58 & 92.96 \\
& 50 & 48.99 & 93.50 \\
& 52.5 & 45.92 & 92.86 \\
\bottomrule
\end{tabular}
\label{tab:sens_thresh_results}
\end{table}

Experiments in Sections \ref{sec:estimator_results} and \ref{sec:main_results} assumed that all filters contributed equally to the information flow downstream and hence, the connectivity scores were the only constraint used for pruning. 
% However, with the use of our sensitivity measure we can highlight critical filters which need to maintain the flow of information to and from them.
% By protecting them from pruning, we can push the extent to which the remaining, relatively insensitive, filters are pruned.
In this section, we highlight the impact of using the sensitivity criterion to prioritize the pruning of relatively weaker filters while protecting more sensitive filters from pruning on the CIFAR10-ResNet56 experimental setup.
In Figs.~\ref{fig:sens_mask1} and \ref{fig:sens_mask2}, we illustrate the 2D pruning masks generated by our algorithm, where the colors black and white represent filters that are removed and retained, respectively, and we observe three distinct behaviours. 
Firstly, when a filter is protected from pruning, an entire row representing all of its associated connections, are retained.
Secondly, in addition to this we also observe an increase in the number of weights pruned from filters that are not protected. 
This is illustrated by an increase in the number of black pixels overall.
Finally, when the sensitivity criterion is applied to layers which were previously not pruned to a large extent (Fig.~\ref{fig:comp_per_layer_figures} Convolution 32, 34, and many others) we observe a complete restructure in the way filters are pruned.
Fig.~\ref{fig:sens_mask2} highlights this trend, which showcases an increase in the overall pruning of the layer as well as a stark difference in how it is pruned.
% All these observations contri
% observation is that for the remaining relatively insensitive set of filters we observe an increase in the number of weight values pruned. 
% illustrate the impact of protecting a few sensitive filters from pruning in convolutional layers 4 and 9 in VGG16. 
All these observations put together lead to an overall improvement in the Pruning~($\%$) with the inclusion of sensitivity, while maintaining high Test Accuracy~($\%$) as shown in Table~\ref{tab:sens_results}.

Across the results presented in Table~\ref{tab:sens_results}, the percentage of filters protected from pruning are maintained at an optimal level.
We determine the optimal combination of high sparsity and accuracy by constraining the $\%$ of filters saved to a value such that SVM model performance is higher than the case when no filters are protected.
The performance comparison is restricted to SVM model only and no re-training is necessary.
When we relax this constraint (Table~\ref{tab:sens_thresh_results}), we observe that the performance levels drop by a significant amount while the sparsity level is lower than expected.
This highlights the necessity of maintaining our constraints in order to obtain the optimal combination of high sparsity with accuracy.

% Results provided in Table~\ref{tab:sens_results} show that by applying sensitivity  we are able to improve the overall compression~($\%$) while maintaining test accuracy.
% \textcolor{red}{Based on the layers we protect where we apply sensitivity, we observe that this approach is \textbf{applicable to layers that are closer to the input and the output}.}

% \textcolor{red}{With the use of a normalization constant across the contributions of each filter, the scale of all the $\lambda$ values is maintained the same, however, the distribution of $\lambda$ values vary depending on the position of the layer.
% Early layers contain $\lambda$ values that mimic a normal distribution while the latter layers do not conform to any specific pattern.}
% Empirical details of the experiment as well as the graphs for $\lambda$ values are provided in the Supplementary Materials.

\section{Conclusion}
Overall, we propose a novel DNN pruning algorithm called \alg{} which uses ACMI to measure the connectivity between filters, a simple set of operating constraints to automate the definition of upper pruning percentage limits of layers in a DNN and a sensitivity criterion that helps protect a subset of critical filters from pruning.
\alg{} provides a faster overall run-time and improves accuracy in the estimation process,  offers state-of-the-art levels of compression using a single train-prune-retrain cycle while the sensitivity criterion can be used to further boost the compression performance.
An important direction of future work is to extend this algorithm to an iterative approach and incorporate it into the training phase.
Doing so would help reduce the overall training time while achieving extreme levels of sparsity.
Additionally, characterizing the pruned networks using a multitude of events like adversarial attacks, calibration error and many others could shed light on how close such networks are to being deployed in the real-world.

% if have a single appendix:
%\appendix[Proof of the Zonklar Equations]
% or
%\appendix  % for no appendix heading
% do not use \section anymore after \appendix, only \section*
% is possibly needed

% use appendices with more than one appendix
% then use \section to start each appendix
% you must declare a \section before using any
% \subsection or using \label (\appendices by itself
% starts a section numbered zero.)
%

\appendices
\section{Bounds on AMI}
Recall the definition of AMI (Eqn.~\ref{def:AMI}). 
For the particular case of $g$, $g(t) = \frac{(t-1)^2}{2(t+1)}$, we have
\begin{align}
\label{def:AMI:arlter}
I_\varphi(X;Y)=&\frac{1}{2}\mathop{\mathbb{E}}_{P_X P_Y}\left[\varphi(X,Y) \left(\frac{dP_{XY}}{dP_X P_Y}+1\right)\right] \\
&- 2 \mathop{\mathbb{E}}_{P_X P_Y}\left[\varphi(X,Y)h\left(\frac{dP_{XY}}{dP_X P_Y}\right)\right], 
\end{align}
where $h(t)=\diy\frac{t}{t+1}$. When $\frac{dP_{XY}}{dP_X P_Y}=1$, then the minimum value of $I_\varphi$ is zero. Further, when $P_{XY}$ and $P_XP_Y$ have no overlapping space then the second term in (\ref{def:AMI:arlter}) becomes zero. Therefore, bounds on $I_\varphi$ is given as,
\begin{equation}\label{bound.AMI}
    0\leq I_\varphi(X,Y)\leq \frac{1}{2}\mathop{\mathbb{E}}_{P_X P_Y}\left[\varphi(X,Y) \left(\frac{dP_{XY}}{dP_X P_Y}+1\right)\right].
\end{equation}

% you can choose not to have a title for an appendix
% if you want by leaving the argument blank
\section{Proof of Theorem 1}
Recall our estimator in Section~\ref{sec:weighted_hash_based_cmi_estimator},  
\begin{equation}
    \widehat{I}_{\varphi}(X;Y|Z) = \sum_{e_{ijk}\in E_G}  {\varphi}(i,j,k)\; \alpha_{ijk}\; g\left(\frac{r_{ijk}}{\alpha_{ijk}}\right), 
\end{equation}
where $\alpha_{ijk}=\diy\frac{r_{ik}\; r_{jk}}{r_k}$. 
The expectation of $\widehat{I}_{\varphi}$ is derived as
\begin{align}
&\bbE\left[\sum_{e_{ijk}\in E_G}  {\varphi}(i,j,k)\; \alpha_{ijk}\; g\left(\frac{r_{ijk}}{\alpha_{ijk}}\right)\big|E_G\right]\\
&\quad=\sum_{e_{ijk}\in E_G}  \bbE\left[{\varphi}(i,j,k)\; \alpha_{ijk}\; g\left(\frac{r_{ijk}}{\alpha_{ijk}}\right)\big|E_{ijk}\right],
\label{eq:1}
\end{align}
where $E_{ijk}$ is the event that there is an edge between the vertices $v_i$, $u_j$, and $\omega_k$ in the dependency graph $G(X,Y,Z)$. 
Let hash function $H_1$ map the N i.i.d points $X_k$, $Y_k$, and $Z_k$ to $\tilde{X}_k$, $\tilde{Y}_k$, and $\tilde{Z}_k$. 
Following the notations used in \cite{Mortezaetal2019}, we denote $E^{=1}_{i}$ be the event that there is exactly one vector from $\tilde{X}_i$ that maps to $v_i$ using $H_2$. 
Similarly, we define $E^{=1}_{j}$ and $E^{=1}_{k}$. 
We denote $E^{=1}_{ijk}:=E^{=1}_{i}\cap E^{=1}_{j}\cap E^{=1}_{k}$ and let $\overline{E^{=1}_{ijk}}$ be the complement set of $E^{=1}_{ijk}$. 

We simplify Eqn.~\ref{eq:1} by splitting it into two parts: without collision and due to collision.
Based on the law of total expectation we have,
\begin{align}
\notag
=&\sum_{e_{ijk}\in E_G} P(E^{=1}_{ijk}|E_{ijk})\\ \notag
&\qquad \bbE\left[{\varphi}(i,j,k)\; \alpha_{ijk}g\left(\frac{r_{ijk}}{\alpha_{ijk}}\right)\big| E^{=1}_{ijk},E_{ijk}\right]\\ \notag
&+\sum_{e_{ijk}\in E_G} P({\overline{E^{=1}_{ijk}}}|E_{ijk})\\
&\qquad \bbE\left[{\varphi}(i,j,k)\; \alpha_{ijk}\; g\left(\frac{r_{ijk}}{\alpha_{ijk}}\right)\big| {\overline{E^{=1}_{ijk}}},E_{ijk}\right]. 
\label{eq:biases}
\end{align}

\noindent\textbf{Step 1 Bias on w/o collision:}
Similar to Lemma 7.3 in \cite{Mortezaetal2019}, we derive, 
\begin{equation}
\label{eq:5}
    P(E^{=1}_{ijk} | E_{ijk})=1-O\left(\frac{1}{\epsilon^d N}\right),\;\;\; d=d_X+d_Y+d_Z. 
\end{equation}
This is because all three $|V|$, $|U$, and $|W|$ are upper bounded by $O(\epsilon^{-d})$. Note that $\epsilon$ is a function of $N$. Additionally from \cite{Mortezaetal2019} we infer the following results:
\begin{equation}
\label{eq:3}
\bbE[\alpha_{ijk}]=\frac{\bbE[r_{ik}]\;\bbE[r_{jk}]}{\bbE[r_k]}+O\left(\sqrt{\frac{1}{N}}\right).
    \end{equation}
Note that (\ref{eq:3}) is implied based on the fact that $\mathbb{V}(\alpha_{ijk})\leq O(1/N)$ which is proved by applying Efron-Stein inequality under assumptions {\bf (A1)} and {\bf (A3)}, similar to arguments in Lemma 7.10 from \cite{Mortezaetal2019}. In addition, we have 
\begin{equation}
\label{eq:4}
\bbE\left[\frac{r_{ijk}}{\alpha_{ijk}}\right]=\frac{\bbE[r_{ijk}]}{\bbE[\alpha_{ijk}]}+O\left(\sqrt{\frac{1}{N}}\right),
\end{equation}

\begin{align}
\label{eq.2}
\notag
\bbE\left[\frac{r_{ijk}}{\alpha_{ijk}}\right]&=P(E^{\leq 1}_{ijk})\;\bbE\left[\frac{r_{ijk}}{\alpha_{ijk}}|E^{\leq 1}_{ijk}\right]\\
&\quad +P(E^{>1}_{ijk}) \bbE\left[\frac{r_{ijk}}{\alpha_{ijk}}|E^{> 1}_{ijk}\right],
\end{align}

where by using similar arguments as in Eqn. 56 from \cite{Mortezaetal2019}, we have $ P(E^{\leq 1}_{ijk}) =1-O(\sqrt{1/(\epsilon^d N)})$.
Therefore, $P(E^{> 1}_{ijk}) =O(\sqrt{1/(\epsilon^d N)})$. 
Further the second term in Eqn.~\ref{eq.2} is the bias because of collision of $H$, which will be proved in the following section, that is upper bounded by $O(\sqrt{1/(\epsilon^d N)})$. 

Let $x_D$ and $x_C$ respectively denote the discrete and continuous components of the vector $x$, with dimensions $d_D$ and $d_C$ . Also let $f_{X_C}(x_C )$ and $p_{X_D} (x_D)$ respectively denote density and pmf functions of these components associated with the probability measure $P_X$. Let $X$ have $d_C$ and $d_D$, $Y$ have $d'_C$,$d'_D$, and $Z$ have $d^{''}_C$, $d^{''}_D$ as their continuous and discrete components, respectively. 
Then it can be shown that,
\begin{align}
\notag
    \bbE[r_{ijk}|E^{\leq 1}_{ijk}] =P(X_D=x_D,Y_D=y_D,Z_D=z_D)\;\\ \epsilon^{d_C+d'_C+d^{''}_C}    \left(f(x_C,y_C,z_C|x_D,y_D,Z_D)+\Delta(\epsilon,q,\gamma)\right),
\end{align}
where densities have bounded derivatives up to the order $q\geq 0$ and belong to the H\"{o}lder continuous class with smoothness parameter $\gamma$. Note that $\Delta(\epsilon,q,\gamma)\rightarrow 0$ as $N\rightarrow \infty$. Now from Eqns. 50, 51, and 53 in \cite{Mortezaetal2019} and from Eqn.~\ref{eq:3},~\ref{eq:4} above, under assumptions {\bf (A1)} and {\bf (A3)}, we derive
\begin{equation} \label{eq:13}
  \bbE\left[\frac{r_{ijk}}{\alpha_{ijk}}|E^{\leq 1}_{ijk}\right]= \frac{dP_{XYZ}\; P_Z}{dP_{XZ} \; P_{YZ}}+\widetilde{\Delta}(\epsilon,q,\gamma)+O\left(\sqrt{\frac{1}{N}}\right),
\end{equation}
where $H(x)=i$, $H(y)=j$, $H(z)=k$, and as $N\rightarrow \infty$,  $\widetilde{\Delta}(\epsilon,q,\gamma)\longrightarrow 0$.

\noindent\textbf{Step 2 Bias because of collision:}
Let $\widetilde{\bX}=\left\{\widetilde{X}_i\right\}_{i=1}^{L_X}$, $\widetilde{\bY}=\left\{\widetilde{Y}_i\right\}_{i=1}^{L_Y}$, $\widetilde{\bZ}=\left\{\widetilde{Z}_i\right\}_{i=1}^{L_Z}$ respectively denote distinct outputs of $H_1$ with the $N$ i.i.d points $X_k$, $Y_k$, $Z_k$ as inputs. 
We denote $L_{XYZ}:=|\widetilde{\bX}\cup\widetilde{\bY}\cup\widetilde{\bZ}|$, $L_{XZ}:=|\widetilde{\bX}\cup\widetilde{\bZ}|$, and $L_{YZ}:=|\widetilde{\bY}\cup\widetilde{\bZ}|$. 
% In this step of the proof, we discuss the bias caused by the collision of $H_1$ that is the second line in (\ref{eq:biases}):
\begin{align}
\label{eq:6}
\notag
\mathbb{B}_\varphi:&=\diy\sum_{e_{ijk}\in E_G} P({\overline{E^{=1}_{ijk}}}|E_{ijk}) \\ \notag
&\qquad \qquad \bbE\left[{\varphi}(i,j,k)\; \alpha_{ijk}\; g\left(\frac{r_{ijk}}{\alpha_{ijk}}\right)\big|
{\overline{E^{=1}_{ijk}}},E_{ijk}\right]\\ \notag
&\leq \diy\sum_{i,j,k\in\mathcal{F}} P(E^{>1}_{ijk}) \\ 
&\qquad \qquad \bbE\left[{\mathds{1}}_{E_{ijk}}{\varphi}(i,j,k)\; \alpha_{ijk}\; g\left(\frac{r_{ijk}}{\alpha_{ijk}}\right)\big|E^{>1}_{ijk}\right],
\end{align}
where 
$E^{>1}_{ijk}= E^{>1}_{i}\cap E^{>1}_{j} \cap E^{>1}_{k}$,
and $E^{>1}_{i}$ is the event that there are at least two vectors from $\tilde{X}_i$ that map to $v_i$ using $H_2$. 
Once again, using the law of total expectation, then the RHS of Eqn.~\ref{eq:6} becomes 
\begin{align}
\label{eq:6-1}
\notag
&=\sum_{i,j,k\in\mathcal{F}} P(E^{>1}_{ijk})\bigg(P(E_{ijk}|E^{>1}_{ijk}) \\ \notag
& \qquad \qquad \bbE\left[{\varphi}(i,j,k)\; \alpha_{ijk}\; g\left(\frac{r_{ijk}}{\alpha_{ijk}}\right)\big|E^{>1}_{ijk},E_{ijk}\right] \\\notag
&\qquad \qquad +P(\overline{E_{ijk}}|E^{>1}_{ijk})\\\notag
& \qquad \qquad \bbE\left[{\varphi}(i,j,k)\; \alpha_{ijk}\; g\left(\frac{r_{ijk}}{\alpha_{ijk}}\right)\big| E^{>1}_{ijk},\overline{E_{ijk}}\right]\bigg)\\\notag
&=\sum_{i,j,k\in\mathcal{F}} P(E_{ijk})P(E^{>1}_{ijk}|E_{ijk})\\
&\qquad\qquad \bbE\left[{\varphi}(i,j,k)\; \alpha_{ijk}\;
g\left(\frac{r_{ijk}}{\alpha_{ijk}}\right)\big|E^{>1}_{ijk},E_{ijk}\right].
\end{align}
The equality in Eqn.~\ref{eq:6-1} is obtained based on Bayes error and $g=0$ on the event $\overline{E_{ijk}}$. 
Now recalling Eqn.~\ref{eq:5}, using Eqn.~\ref{bound.AMI} we bound the last line in Eqn.~\ref{eq:6-1} by, 
\begin{align}
\label{eq:v2-4}
\notag
&\diy O\left(\frac{1}{\epsilon^d N}\right)\sum_{i,j,k\in\mathcal{F}} P(E_{ijk}) \\
&\qquad \qquad \qquad \qquad  \bbE\left[{\varphi}(i,j,k) \left(r_{ijk}+\alpha_{ijk}\right)\big| E^{>1}_{ijk},E_{ijk}\right].
\end{align}

This implies that
\begin{align}
\notag
\mathbb{B}_\varphi &\leq O\left(\frac{1}{\epsilon^d N}\right)\sum_{i,j,k\in\mathcal{F}} P(E_{ijk})\\ \notag
&\qquad \qquad \qquad \qquad   \left(\bbE\left[{\varphi}(i,j,k) r_{ijk}\big|E^{>1}_{ijk},
E_{ijk}\right]\right.\\\notag
&\left. \qquad \qquad \qquad \qquad +\bbE\left[{\varphi}(i,j,k)\alpha_{ijk} \big|E^{>1}_{ijk},E_{ijk}\right]\right)\\ \notag
&=O\left(\frac{1}{\epsilon^d N^2}\right)\sum_{i,j,k\in\mathcal{F}} P(E_{ijk})\\\notag
& \qquad \qquad \qquad \qquad \left( \bbE\left[{\varphi}(i,j,k)\; N_{ijk}\big| E^{>1}_{ijk} ,E_{ijk}\right] \right.\\
& \qquad \qquad \qquad \qquad \left.+\bbE\left[{\varphi}(i,j,k)\frac{N_{ik}N_{jk}}{N_k} \big|E^{>1}_{ijk},E_{ijk}\right]\right).
\end{align}
If we extend our discussion to all the possible mappings from $H_1$ we obtain,
\begin{align}
\notag
&=O\left(\frac{1}{\epsilon^d N^2}\right)\sum_{\tilde{\bx},\tilde{\by},\tilde{\bz}} p_{\tilde{\bX},\tilde{\bY},\tilde{\bZ}}(\tilde{\bx},\tilde{\by},\tilde{\bz}) \sum_{i,j,k\in\mathcal{F}} P(E_{ijk})  \\ \notag
& \qquad \bigg(\bbE\left[{\varphi}(i,j,k)\; N_{ijk}\big|E^{>1}_{ijk},E_{ijk},\tilde{\bX}=\tilde{\bx},\tilde{\bY}=\tilde{\by}, 
\right.\\ \notag 
&\qquad \qquad \left. \tilde{\bZ}=\tilde{\bz}\right] \\ \notag
&\qquad +\diy\bbE\left[{\varphi}(i,j,k)\frac{N_{ik}N_{jk}}{N_k} \big| 
E^{>1}_{ijk},E_{ijk},\tilde{\bX}=\tilde{\bx},\tilde{\bY}=\tilde{\by} , \right. \\
&\qquad \qquad \left. \tilde{\bZ}=\tilde{\bz}\right]\bigg).
\label{eq:v2-4-1}
\end{align}

Let us define,
\begin{align}
\notag
\mathcal{A}_{ijk}&:=\left\{r: H_2(\tilde{X}_r)=i, H_2(\tilde{Y}_r)=j,H_2(\tilde{Z}_r)=k\right\}, \\ \notag
\mathcal{A}_{k}&:=\left\{r: H_2(\tilde{Z}_r)=k\right\}, \\ \notag
 \mathcal{A}_{ik}&:=\left\{r: H_2(\tilde{X}_r)=i,H_2(\tilde{Z}_r)=k\right\}, \\
 \mathcal{A}_{jk}&:=\left\{r: H_2(\tilde{Y}_r)=j,H_2(\tilde{Z}_r)=k\right\}. 
\end{align}

Let $M_r$, be the number of the input points $(\bX,\bY,\bZ)$ mapped to $(\tilde{X}_r,\tilde{Y}_r,\tilde{Z}_r)$. Therefore for $i,j,k$ we can rewrite $N_{ijk}$ as 
\begin{equation}
    N_{ijk}=\diy\sum_{r=1}^{L_{XYZ}} {\mathds{1}}_{A_{ijk}}(r)M_r.
\end{equation}
Similarly $M'_r$, $\widetilde{M}_s$, and $\overline{M}_t$ are defined the number of the input points mapped to $(\tilde{X}_r,\tilde{Z}_r)$, $(\tilde{Y}_s,\tilde{Z}_s)$, and $\tilde{Z}_t$, respectively and we can write 
\begin{align}
     &N_{ik}=\diy\sum_{r=1}^{L_{XZ}} {\mathds{1}}_{A_{ik}}(r)M'_r, \;\; \;\;   N_{jk}=\diy\sum_{s=1}^{L_{YZ}} {\mathds{1}}_{A_{jk}}(s)\widetilde{M}_s, \\ 
     &\qquad \qquad \qquad \qquad  N_{k}=\diy\sum_{t=1}^{L_{Z}} {\mathds{1}}_{A_{k}}(t)\overline{M}_t.
\end{align}

Under the assumption that $\varphi$ is bounded, we have
\begin{align}
\label{eq:v2-6}
\notag
\mathbb{B}_\varphi &\leq \diy O\left(\frac{1}{\epsilon^d N^2}\right) \diy\sum_{\tilde{\bx},\tilde{\by},\tilde{\bz}} p_{\tilde{\bX},\tilde{\bY},\tilde{\bZ}}(\tilde{\bx},\tilde{\by},\tilde{\bz}) \diy\sum_{i,j,k\in\mathcal{F}} P(E_{ijk}) \\ \notag
&\qquad \bigg(\diy\sum_{r=1}^{L_{XYZ}}P\left(r\in \mathcal{A}_{ijk}\big|E^{>1}_{ijk},E_{ijk},\tilde{\bX}=\tilde{\bx}, \tilde{\bY}=\tilde{\by},\right. \\ \notag
&\qquad \left. \tilde{\bZ}=\tilde{\bz}\right) \bbE\left[ M_r\big|E^{>1}_{ijk},E_{ijk},\tilde{\bX}=\tilde{\bx}, \tilde{\bY}=\tilde{\by},\tilde{\bZ}=\tilde{\bz}\right]\\ \notag
&\qquad+\diy\sum_{r=1}^{L_{XZ}}\sum_{s=1}^{L_{YZ}}\sum_{t=1}^{L_Z}P\left(r\in \mathcal{A}_{ik},s\in \mathcal{A}_{jk},t\in\mathcal{A}_{k}\big|E^{>1}_{ijk},\right.\\ \notag
&\quad \qquad \left. E_{ijk},\tilde{\bX}=\tilde{\bx},  \tilde{\bY}=\tilde{\by},\tilde{\bZ}=\tilde{\bz}\right) \\
&\quad \qquad \diy\bbE\left[\frac{M'_{r}\widetilde{M}_{s}}{\overline{M}_t} \big|E^{>1}_{ijk}, E_{ijk},\tilde{\bX}=\tilde{\bx}, \tilde{\bY}=\tilde{\by},\tilde{\bZ}=\tilde{\bz}\right]\bigg). 
\end{align}

Next we find the probability terms:
\begin{equation}
\label{eq:10}
\begin{array}{l}
P\left(r\in \mathcal{A}_{ijk}\big|E^{>1}_{ijk},E_{ijk},\tilde{\bX}=\tilde{\bx}, \tilde{\bY}=\tilde{\by},\tilde{\bZ}=\tilde{\bz}\right)\\[10pt]
=\diy\frac{P\left(r\in \mathcal{A}_{ijk},E^{>1}_{ijk} |\tilde{\bX}=\tilde{\bx}, \tilde{\bY}=\tilde{\by},\tilde{\bZ}=\tilde{\bz}\right)}{P\left(E^{>1}_{ijk}|\tilde{\bX}=\tilde{\bx}, \tilde{\bY}=\tilde{\by},\tilde{\bZ}=\tilde{\bz}\right)}. 
\end{array}
\end{equation}
We first find the denominator of Eqn.~\ref{eq:10} first. 
We define $a=1$ when $i=j=k$ and $a=3$ for the case $i\neq j\neq k$:
\begin{align}
\label{eq:v2-1}
\notag
&P\left(E^{>1}_{ijk}|\tilde{\bX}=\tilde{\bx}, \tilde{\bY}=\tilde{\by},\tilde{\bZ}=\tilde{\bz}\right)\\\notag
&=1-P\left(E^{=0}_{ijk}|\tilde{\bX}=\tilde{\bx}, \tilde{\bY}=\tilde{\by},\tilde{\bZ}=\tilde{\bz}\right)\\\notag
&\qquad -P\left(E^{=1}_{ijk}|\tilde{\bX}=\tilde{\bx}, \tilde{\bY}=\tilde{\by},\tilde{\bZ}=\tilde{\bz}\right)\\\notag
&= 1-\left(\diy\frac{F-a}{F}\right)^{ L_{XYZ}}-\left(\diy\frac{L_{XYZ}}{F^a}\left(\diy\frac{F-a}{F}\right)^{L_{XYZ}-a}\right)\\
&=O\left(\diy\frac{L_{XYZ}^2}{F^{a+1}}\right).
\end{align}

Further, 
 \begin{align}
 \label{eq:V2-2}
 \notag
 &P\left(r\in \mathcal{A}_{ijk},E^{>1}_{ijk} |\tilde{\bX}=\tilde{\bx}, \tilde{\bY}=\tilde{\by},\tilde{\bZ}=\tilde{\bz}\right) \\\notag
&= P\left(r\in \mathcal{A}_{ijk} |E^{>1}_{ijk},\tilde{\bX}=\tilde{\bx}, \tilde{\bY}=\tilde{\by},\tilde{\bZ}=\tilde{\bz}\right) \\\notag
&\qquad P\left(r\in \mathcal{A}_{ijk}|\tilde{\bX}=\tilde{\bx}, \tilde{\bY}=\tilde{\by},\tilde{\bZ}=\tilde{\bz}\right) \\
&=\left(1-\left(\diy\frac{F-a}{F}\right)^{L_{XYZ}-a}\right)\diy\left(\frac{1}{F}\right)^a=O\left(\frac{L_{XYZ}}{F^{a+1}}\right).
 \end{align}
 
Combining Eqn.~\ref{eq:v2-1} and \ref{eq:V2-2} yields 
 \begin{align}
 \label{eq:v2-4}
 \notag
&P\left(r\in \mathcal{A}_{ijk}\big|E^{>1}_{ijk},E_{ijk},\tilde{\bX}=\tilde{\bx}, \tilde{\bY}=\tilde{\by},\tilde{\bZ}=\tilde{\bz}\right)\\
&=\quad O\left(\diy\frac{1}{L_{XYZ}}\right).
 \end{align}
 
 Now we simplify the following term:
 \begin{align}
 \notag
     &P\left(r\in \mathcal{A}_{ik},s\in \mathcal{A}_{jk},t\in\mathcal{A}_{k}\big|E^{>1}_{ijk},E_{ijk},\right. \\
      &\qquad \qquad \qquad \left. \tilde{\bX}=\tilde{\bx}, \tilde{\bY}=\tilde{\by},\tilde{\bZ}=\tilde{\bz}\right).
 \end{align}
 
First we assume that $\tilde{X}_v\neq \tilde{Y}_v\neq \tilde{Z}_v$ for $v=r,s,t$. Then
\begin{align}
\label{eq:v2-3}
\notag
&P\left(r\in \mathcal{A}_{ik},s\in \mathcal{A}_{jk},t\in\mathcal{A}_{k}\big|E^{>1}_{ijk},E_{ijk},\tilde{\bX}=\tilde{\bx}, \tilde{\bY}=\tilde{\by},\right. \\ \notag
&\quad \quad \left.\tilde{\bZ}=\tilde{\bz}\right) \\ \notag
&\leq P\left(r\in \mathcal{A}_{ik}\big|E^{>1}_{ik},\tilde{\bX}=\tilde{\bx}, \tilde{\bY}=\tilde{\by},\tilde{\bZ}=\tilde{\bz}\right) \\ \notag
&\qquad P\left(s\in \mathcal{A}_{jk}\big|E^{>1}_{jk},\tilde{\bX}=\tilde{\bx}, \tilde{\bY}=\tilde{\by},\tilde{\bZ}=\tilde{\bz}\right)\\ \notag
&\qquad P\left(t\in\mathcal{A}_{k}\big|E^{>1}_{k},\tilde{\bX}=\tilde{\bx}, \tilde{\bY}=\tilde{\by},\tilde{\bZ}=\tilde{\bz}\right)\\
&=O\left(\diy\frac{1}{L_{XZ}L_{YZ}L_{Z}}\right). 
 \end{align}
 
 Next assume that $\tilde{X}_v= \tilde{Y}_v= \tilde{Z}_v$ for $v=r,s,t$, therefore $H_2(\tilde{X}_v)=H_2( \tilde{Y}_v)= H_2(\tilde{Z}_v)$, for $v=r,s,t$. Then 
 \begin{equation}
 \label{eq:v2-2}
 \begin{array}{l}
     P\left(r\in \mathcal{A}_{ik},s\in \mathcal{A}_{jk},t\in\mathcal{A}_{k}\big|E^{>1}_{ijk},E_{ijk},\tilde{\bX}=\tilde{\bx}, \tilde{\bY}=\tilde{\by},\right.\\
     \qquad \left. \tilde{\bZ}=\tilde{\bz}\right) =\delta_{ijk} O\left(\diy\frac{1}{L_{XYZ}}\right). 
     \end{array}
 \end{equation}
 \def\bw{\mathbf{w}}
%where   $L_{XYZ}:=|\tilde{\bw}|$, $\tilde{\bw}=\tilde{\bx}\cup\tilde{\by}\cup \tilde{\bz}$. 
By using Eqns.~\ref{eq:v2-2}, \ref{eq:v2-3}, and \ref{eq:v2-4} in Eqn.~\ref{eq:v2-6} we obtain an upper bound on bias with collision:

\begin{align}
\label{eq:v2-7}
\notag
\mathbb{B}_\varphi&\leq \diy O\left(\frac{1}{\epsilon^d N^2}\right) \diy\sum_{\tilde{\bx},\tilde{\by},\tilde{\bz}} p_{\tilde{\bX},\tilde{\bY},\tilde{\bZ}}(\tilde{\bx},\tilde{\by},\tilde{\bz}) \diy\sum_{i,j,k\in\mathcal{F}} P(E_{ijk}) \\\notag
&\quad \bigg(O\left(\diy\frac{1}{L_{XYZ}}\right)\diy\sum_{r=1}^{L_{XYZ}}\bbE\left[ M_r\big|E^{>1}_{ijk},E_{ijk},\tilde{\bX}=\tilde{\bx}, \right.\\ \notag
&\qquad \qquad \qquad \qquad  \qquad \quad \left. \tilde{\bY}=\tilde{\by},\tilde{\bZ}=\tilde{\bz}\right]\\\notag 
&\quad +\diy\left(O\left(\diy\frac{1}{L_{XZ}L_{YZ}L_{Z}}\right)+\delta_{ijk}O\left(\diy\frac{1}{L_{XYZ}}\right)\right) \\\notag
&\qquad \diy\sum_{r=1}^{L_{XZ}}\sum_{s=1}^{L_{YZ}}\sum_{t=1}^{L_Z}
\diy\bbE\left[\frac{M'_{r}\widetilde{M}_{s}}{\overline{M}_t} \big|E^{>1}_{ijk}, 
 E_{ijk},\tilde{\bX}=\tilde{\bx}, \right. \\ \notag
&\qquad \qquad \qquad \qquad \quad \left. \tilde{\bY}=\tilde{\by},\tilde{\bZ}=\tilde{\bz}\right]\bigg)\\\notag
&= \diy O\left(\frac{1}{\epsilon^d N^2}\right) \diy\sum_{\tilde{\bx},\tilde{\by},\tilde{\bz}} p_{\tilde{\bX},\tilde{\bY},\tilde{\bZ}}(\tilde{\bx},\tilde{\by},\tilde{\bz}) \diy\sum_{i,j,k\in\mathcal{F}} P(E_{ijk})\\ \notag
&\quad \bigg(O\left(\diy\frac{N}{L_{XYZ}}\right)+\diy\left(O\left(\diy\frac{N}{L_{XZ}L_{YZ}L_{Z}}\right)+\right. \\
&\quad \left. \delta_{ijk}O\left(\diy\frac{N}{L_{XYZ}}\right)\right)\bigg).
\end{align}
Re-arranging the expectation term we get,
\begin{align}
\notag
&= \diy O\left(\frac{1}{\epsilon^d N^2}\right) \diy\sum_{\tilde{\bx},\tilde{\by},\tilde{\bz}} p_{\tilde{\bX},\tilde{\bY},\tilde{\bZ}}(\tilde{\bx},\tilde{\by},\tilde{\bz}) \bigg(O\left(\diy\frac{N}{L_{XYZ}}\right)+\\ \notag
&\qquad \diy\left(O\left(\diy\frac{N}{L_{XZ}L_{YZ}L_{Z}}\right)+O\left(\diy\frac{1}{L_{XYZ}}\right)\right) \bigg)\\ \notag
&\qquad \bbE\left[ \diy\sum_{i,j,k\in\mathcal{F}}{\mathds{1}}_{E_{ijk}}\right]\\ \notag
&\leq \diy O\left(\frac{1}{\epsilon^d N^2}\right) \diy\sum_{\tilde{\bx},\tilde{\by},\tilde{\bz}} p_{\tilde{\bX},\tilde{\bY},\tilde{\bZ}}(\tilde{\bx},\tilde{\by},\tilde{\bz}) \bigg(O\left(\diy\frac{N}{L_{XYZ}}\right)+\\ \notag
&\qquad \diy\left(O\left(\diy\frac{N}{L_{XZ}L_{YZ}L_{Z}}\right)+O\left(\diy\frac{1}{L_{XYZ}}\right)\right)\bigg) L_{XYZ}\\
&\leq \diy O\left(\frac{1}{\epsilon^d N}\right).
\end{align}
Hence as $N\longrightarrow \infty$, the bias estimator due to collision tends to zero i.e. $\mathbb{B}_\varphi\longrightarrow 0$. \\

\noindent\textbf{Step 3 Combine Results:}
Let us denote $N'_{ijk}$, $N'_{ik}$, $N'_{jk}$, and $N'_k$ respectively as the number of the input points $(\bX,\bY,\bZ)$, $(\bX,\bZ)$, $(\bY,\bZ)$, and $\bZ$ mapped to the bins $(\tilde{X}_i,\tilde{Y}_j,\tilde{Z}_k)$, $(\tilde{X}_i,\tilde{Z}_k)$, $(\tilde{Y}_j,\tilde{Z}_k)$, and $\tilde{Z}_k$ using $H_1$. 
We define the notations $r(i)=H^{-1}_2(i)$ for $i\in\mathcal{F}$ and $s(x):=H_1(x)$ for $x\in\mathcal{X}\cup\mathcal{Y}\cup\mathcal{Z}$. Then from Eqn.~\ref{eq:13}, we have
\begin{align} \label{eq:14}
\notag
  &\bbE\left[\diy\frac{N'_{s(X)s(Y)s(Z)}N'_{s(Z)}}{N'_{s(X)s(Z)}N'_{s(Y)s(Z)}}\right] \\
  &=\quad \diy\frac{dP_{XYZ}\; P_Z}{dP_{XZ} \; P_{YZ}}+\widetilde{\Delta}(\epsilon,q,\gamma)+O\left(\sqrt{\frac{1}{N}}\right).
\end{align}

We simplify the first term in Eqn.~\ref{eq:biases} as,
\begin{align}
\label{eq:11}
\notag
&\diy\sum_{i,j,k\in\mathcal{F}} P(E^{\leq 1}_{ijk}) \bbE\left[{\mathds{1}}_{E_{ijk}}{\varphi}(i,j,k)\; \alpha_{ijk}\;
g\left(\frac{r_{ijk}}{\alpha_{ijk}}\right)\big|E^{\leq1}_{ijk}\right]\\\notag
&=\diy\left(1-O\left(\frac{1}{\epsilon^d N}\right)\right)\diy\sum_{i,j,k\in\mathcal{F}}\bbE\left[{\mathds{1}}_{E_{ijk}}{\varphi}(i,j,k)\; \alpha_{ijk}\; \right.\\\notag
&\qquad \qquad \qquad \qquad \qquad \qquad    \left. g\left(\frac{r_{ijk}}{\alpha_{ijk}}\right)\big|E^{\leq1}_{ijk}\right]\\\notag
&=\diy \diy\sum_{i,j,k\in\mathcal{F}}\bbE\left[{\mathds{1}}_{E_{ijk}}{\varphi}(i,j,k)\; \frac{N_{ik}N_{jk}}{N_k N}\; g\left(\frac{N_{ijk}N_k}{N_{ik}N_{jk}}\right)\big|E^{\leq1}_{ijk}\right] \\ \notag
&\qquad \qquad  + O\left(\frac{1}{\epsilon^d N}\right)\\ \notag
&=\diy \diy\sum_{i,j,k\in\mathcal{F}}\bbE\left[{\mathds{1}}_{E_{ijk}}{\varphi}(r(i),r(j),r(k))\; \frac{N'_{r(i)r(k)}N'_{r(j)r(k)}}{N'_{r(k)} N}\;\right.\\
&\qquad \qquad \qquad  \left. g\left(\frac{N'_{r(i)r(j)r(k)}N'_{r(k)}}{N'_{r(i)r(k)}N'_{r(j)r(k)}}\right)\right] + O\left(\frac{1}{\epsilon^d N}\right).
\end{align}

Lets denote 
$$\beta(r(i),r(j),r(k))=\diy\frac{N'_{r(i)r(j)r(k)}N'_{r(k)}}{N'_{r(i)r(k)}N'_{r(j)r(k)}}.$$
Therefore the last line in Eqn.~\ref{eq:11} is equal to

\begin{align}
\label{eq:12}
\notag
&=\diy\frac{1}{N}\diy \diy\sum_{i,j,k\in\mathcal{F}}\bbE\left[{\varphi}(r(i),r(j),r(k))\; \frac{N'_{r(i)r(j)r(k)}}{\beta(r(i),r(j),r(k))}\;\right.\\ \notag
&\qquad \qquad \qquad \qquad  \left. g\Big(\beta(r(i),r(j),r(k))\Big)\right] + O\left(\frac{1}{\epsilon^d N}\right)\\ \notag
&=\diy\frac{1}{N}\bbE\left[ \diy\sum_{i=1}^N \frac{{\varphi}(s(X),s(Y),s(Z))}{\beta(s(X),s(Y),s(Z))}\; g\Big(\beta(s(X),s(Y),s(Z))\Big)\right]\\
&\qquad + O\left(\frac{1}{\epsilon^d N}\right), 
\end{align}
where 
$$\beta(s(X),s(Y),s(Z))=\diy\frac{N'_{s(X)s(Y)s(Z)}N'_{s(Z)}}{N'_{s(X)s(Z)}N'_{s(Y)s(Z)}}.$$

The expression in Eqn.~\ref{eq:12} equals:
\begin{align}
\label{eq:14}
\notag
&=\diy\bbE_{P_{XYZ}}\left[\bbE\left[ \frac{{\varphi}(s(X),s(Y),s(Z))}{\beta(s(X),s(Y),s(Z))}\; g\Big(\beta(s(X),s(Y)\right. \right.\\ \notag
&\qquad \left. \left. ,s(Z))\Big)\Big|\bX=\bx,\bY=\by,\bZ=\bz\right]\right] + O\left(\frac{1}{\epsilon^d N}\right)\\ \notag
&= \diy\bbE_{P_{XYZ}}\left[{\varphi}(X,Y,Z) h\Big(\diy\frac{dP_{XYZ}\; P_Z}{dP_{XZ} \; P_{YZ}}\Big)\right] +\widetilde{\Delta}(\epsilon,q,\gamma) + \\ 
& \qquad O\left(\sqrt{\frac{1}{N}}\right)+ O\left(\frac{1}{\epsilon^d N}\right),
\end{align}

where $h(t)=g(t)/t$ and Eqn.~\ref{eq:14} is derived by borrowing Lemma 7.9 from \cite{Mortezaetal2019}. Hence from Eqn.~\ref{eq:14} and Eqn.~\ref{eq:biases}, and the fact that $\widetilde{\Delta}(\epsilon,q,\gamma)\longrightarrow 0$ as $N\rightarrow \infty$, we conclude
\begin{equation}
\begin{array}{ll}
\bbE\left[\widehat{I}_{\varphi}(X;Y|Z)\right] \longrightarrow \diy\bbE_{P_{XYZ}}\left[{\varphi}(X,Y,Z) h\Big(\diy\frac{dP_{XYZ}\; P_Z}{dP_{XZ} \; P_{YZ}}\Big)\right], \;\;\\
\hbox{as}\; \;\; N\rightarrow \infty.  
\end{array}
\end{equation}
This completes the proof.

\section{Complexity of \alg{}}
We breakdown the discussion on the computational complexity of \alg{} into two parts,
1) the complexity of the hash-based ACMI estimator, and 2) the complexity of Algorithm 1 in the main paper.

\subsection{Complexity of hash-based estimator}
By extending the discussion provided in \cite{Mortezaetal2019}, we find that the estimation process is dependent on two main factors, the total number of samples, $N$, and the dimensionality of each sample.
From the original paper, we find that the computational complexity is linearly dependent on the number of samples as well as the dimensionality of the samples. 
In our setup the dimensionality of a sample is capped by  $\overline{F^{(l)}_j}$ which includes activations from all the filters in a layer excluding $j$. 
The exact value of this variable is dependent on the neural network architecture over which ACMI is calculated.

\subsection{Complexity of Algorithm 1 (Main Paper)}
There are 2 primary factors which affect the complexity of Algorithm 1 in the main paper, 1) the number of groups associated with each layer $l$ and $l+1$, and 2) the total number of layers in the DNN.
The internal double \textbf{FOR} loop has an upper bound of $\mathcal{O}(N^{(l)}N^{(l+1)})$ if the number of groups defined matches the number of filters in each layer.
% Practically, we have an upper bound of $\mathcal{O}(G^{(l)}G^{(l+1)})$.
The outer \textbf{FOR} loop, used to iterate over pairs of adjacent layers, is executed a total of $L-1$ times.

\section{Validating the Estimator}
In this section we validate the MSE performance of the ACMI estimator across various dimensionalities and total number of samples to asses the trends in estimation accuracy. 

\begin{figure}[t!]
    \centering
    \subfloat[][MSE vs. No. of Samples]{\label{fig:est_samples}\includegraphics[width=\columnwidth]{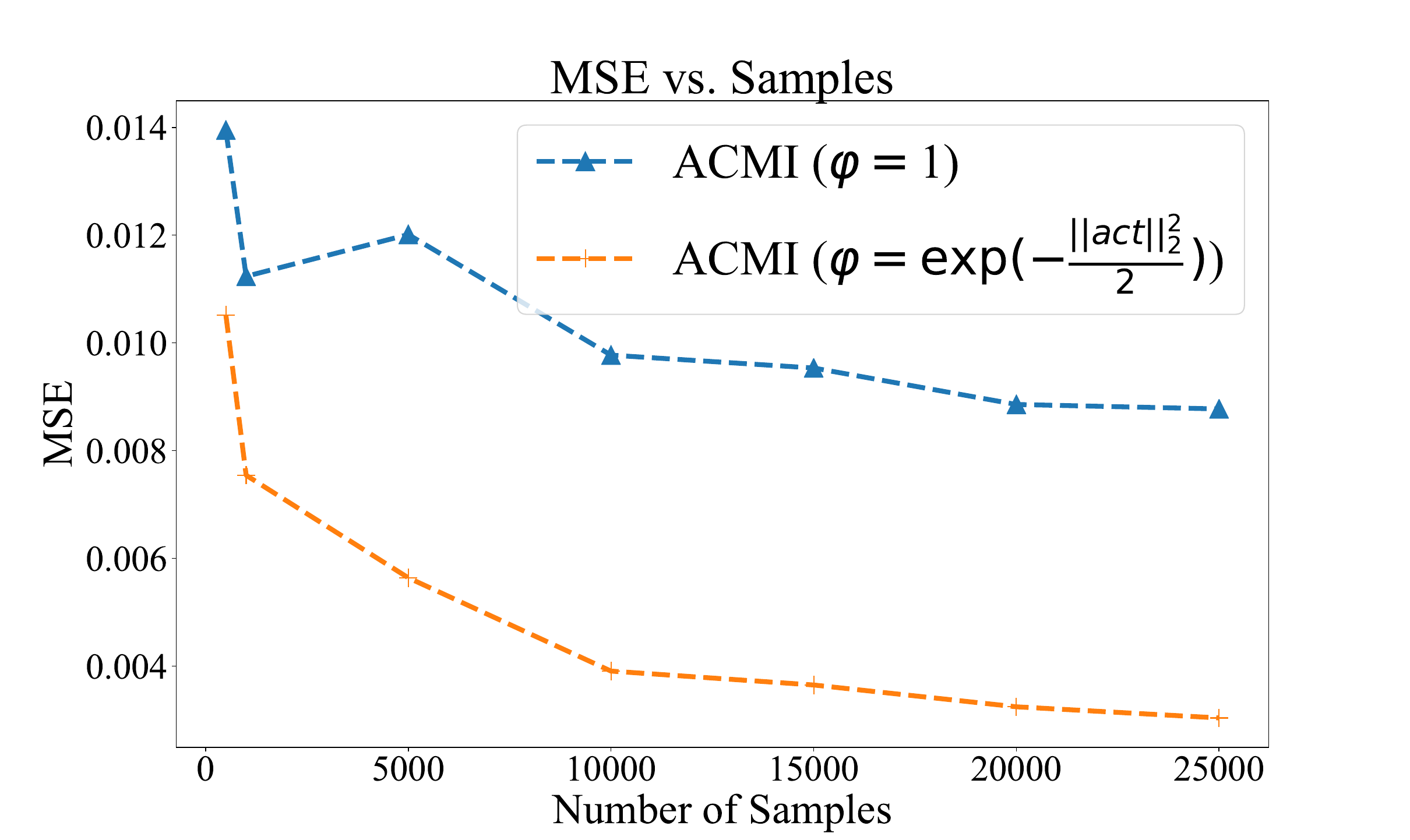}}\\
    \subfloat[][MSE vs. Dimensionality]{\label{fig:est_dims}\includegraphics[width=\columnwidth]{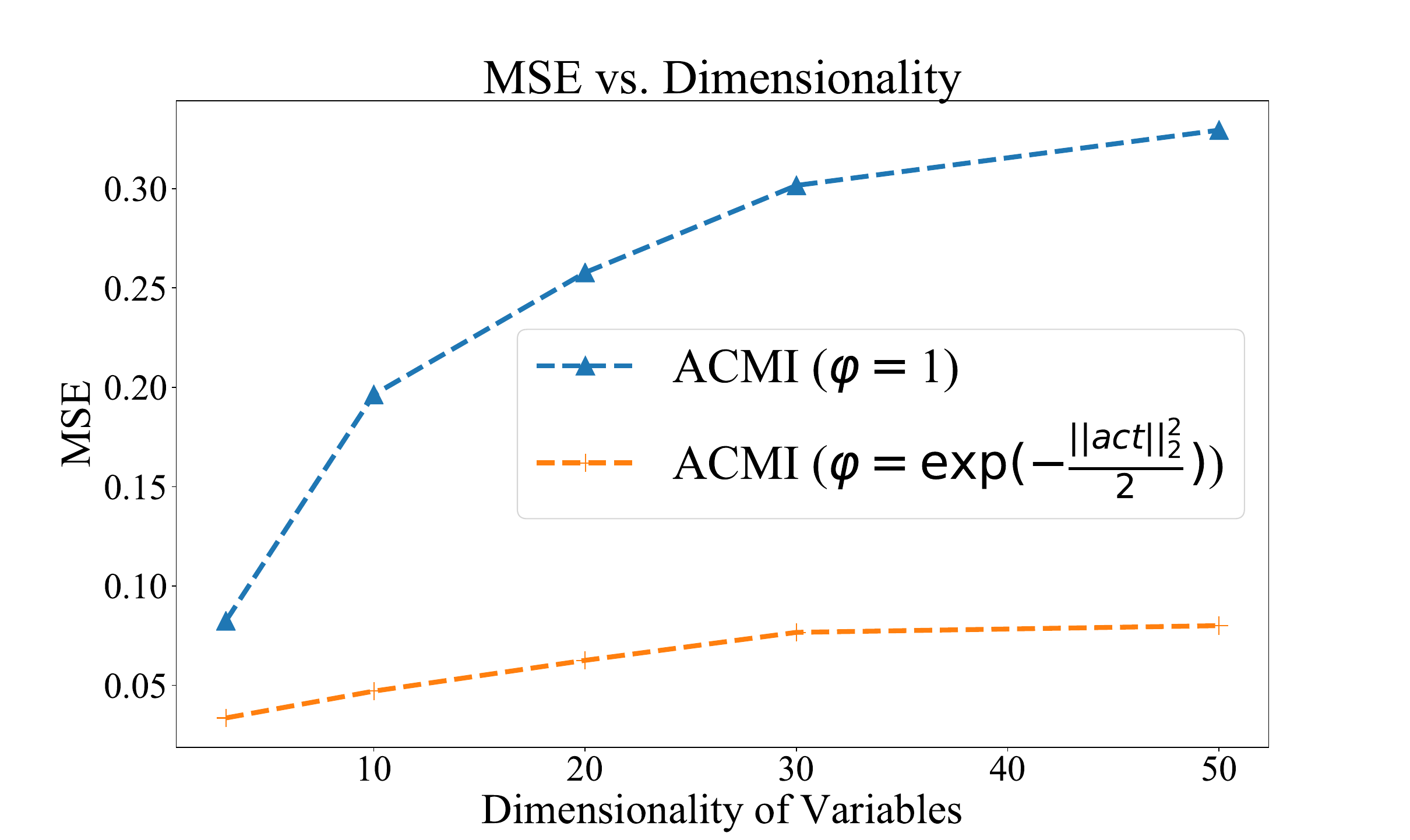}}
    \caption{(\textbf{Fig.~\ref{fig:est_samples}}) An increase in the number of samples while dimensionality of input variables are held constant shows steadily decreasing MSE. (\textbf{Fig.~\ref{fig:est_dims}}) Increasing the dimensionality of input variables while the total number of samples are constant shows a steady decline of the MSE. Overall, the trends observed in both experiments match the expectations from a valid estimator.}
    \label{fig:estimator_validation}
\end{figure}

\subsection{Setup}
To observe the performance of the estimator when the number of samples are varied, we set the dimensionality of $X,Y$ to one and $Z$ to two. 
This setup is used to mimic the dimensionality difference, at a small scale, in our experiments.
We vary the number of samples in the range $\in \{500, 1000, 5000, 10000, 15000, 20000, 25000 \}$.
To observe the impact of a change in dimensionality on the estimator's performance, we restrict the total number of samples to $5000$ and vary the dimensions of $X,Y,Z$ across $\{3, 10, 20, 30, 50\}$.
In both the setups, we sample data from a multivariate normal distribution where the covariance matrix is set as the identity function and $\mu$ is zero.

\subsection{Results}
Fig.~\ref{fig:estimator_validation} shows the results of our experiments where in Fig.~\ref{fig:est_samples}, we observe the steady decrease in MSE as the number of samples are increased. 
This matches our expectation of a good estimator where an increase in the number of samples improves the overall estimation accuracy and thus, reduces the MSE.
Fig.~\ref{fig:est_dims} illustrates the steady increase in MSE when the number of samples are held constant but the dimensionality of the input variables grows larger.
Further, the trends from secondary curves with $\varphi=\exp(-\frac{\norm{\textrm{act}}_2^2}{2})$ show that the inclusion of a scaling term improves the overall performance.
Thus, our observations match the expected trends from a valid estimator.

\section{Dataset and Preprocessing}
\subsection{CIFAR10}
This dataset is a 10 class subset of the original 80 million tiny images dataset.
The dataset split contains 50000 images for training, split as 5000 images/class, and 10000 images for testing where there are 1000 images/class.
Each image in the dataset is originally $32\times 32 \times 3$.
For preprocessing, we randomly crop the image after padding 4 pixels, then we randomly flip the image horizontally before normalizing its values using mean (0.4914, 0.4822, 0.4465) and std. (0.2470, 0.2435, 0.2616) for each channel respectively.
During testing, the images are only normalized and provided to the DNN.
\subsection{ILSVRC2012}
This dataset contains 1000 different classes of images totalling to about 1.2 million images overall for training and 50000 images for validation.
The number of images per class varies between 732 to 1300. 
For preprocessing, we randomly crop the image in to $224 \times 224 \times 3$, then we randomly flip the image horizontally before normalizing its values using mean (0.485, 0.456, 0.406) and std. (0.229, 0.224, 0.225) for each channel respectively.
During testing, we resize the original image to $256 \times 256 \times 3$, take a center crop of size $224 \times 224 \times 3$ before normalizing it and providing it to the DNN.

\section{Experimental Setup}
Throughout our experiments we use three major Dataset-DNN combinations, CIFAR10-VGG16, CIFAR10-ResNet56 and ILSVRC2012-ResNet50.
Table~\ref{tab:training_setups} lists the main hyper-parameters used to train the VGG16 and ResNet56 networks and obtain their baseline performances.
Pre-trained weights for ILSVRC2012-ResNet50 are used to compute ACMI values.
Table~\ref{tab:base_retraining} list the basic hyper-parameters used to retrain the VGG16, ResNet56 and ResNet50 networks and obtain their final performance.

\begin{table}[h]
\centering
\caption{Training setups used to obtain pre-trained network weights}
\begin{tabular}{@{}lcc@{}}
\toprule
              & VGG16  & ResNet56  \\ \midrule
Epochs        & 300             & 300\\
Batch Size    & 128             & 128\\
Learning Rate & 0.1             & 0.01\\
Schedule      & 90, 180, 260    & 150, 225\\
Optimizer     & SGD             & SGD\\
Weight Decay  & 0.0005          & 0.0002\\
Multiplier    & 0.2             & 0.1\\
\bottomrule
\end{tabular}
\label{tab:training_setups}
\end{table}

\begin{table}[h]
\centering
\caption{Base retraining setup used to obtain final performance listed in Table 1 of main paper}
\begin{tabular}{@{}lccc@{}}
\toprule
       & VGG16  & ResNet56 & ResNet50 \\ \midrule
Epochs         & 300             & 300             & 100 \\
Batch Size     & 128             & 128             & 64 \\
Learning Rate  & 0.1             & 0.1             & 0.1  \\
Schedule       & [90, 180, 260]  & [90, 180, 260]  & [30, 60, 90]  \\
Optimizer      & SGD             & SGD             & SGD  \\
Weight Decay   & 0.0005          & 0.0005          & 0.0001/0.00003 \\
Multiplier     & 0.1             & 0.2             & 0.1  \\
Label Smoothing& 0.35            & 0.15            & 0.9  \\
\bottomrule
\end{tabular}
\label{tab:base_retraining}
\end{table}
\begin{table*}[ht!]
\centering
\caption{Hyper-parameters specific to the $\varphi$ function used final performance the best possible final performance $\geq 93.43\%$. Here, $\textrm{act}$ refers to the activations and $\gamma$ values are represented as $\%$}
\begin{tabular}{@{}lcccccccc@{}}
\toprule
               & $1$ & $\norm{\textrm{weights}}_2$ & $\norm{\textrm{weights}}_2^2$ & $\exp(\frac{-\norm{\textrm{weights}}_2^2}{2})$ & $\norm{\textrm{act}}_2$ & $\norm{\textrm{weights}}_2\norm{\textrm{act}}_2$  & $\exp(-\frac{\norm{\textrm{weights}}_2^2\norm{\textrm{act}}_2^2}{2})$\\ \midrule\\
$\delta$       & 0.9865  & 0.9925 & 0.9925 & 0.988 & 0.995  & 0.880   & 0.919\\  
$\gamma^{(1)}$ & 00.00   & 00.00  & 00.00  & 00.00 & 00.00  & 00.00   & 00.00\\  
$\gamma^{(2)}$ & 00.00   & 00.00  & 00.00  & 00.00 & 00.00  & 00.00   & 00.00\\  
$\gamma^{(3)}$ & 21.02   & 21.02  & 21.02  & 21.02 & 00.00  & 41.01   & 36.03\\  
$\gamma^{(4)}$ & 51.02   & 51.02  & 51.02  & 51.02 & 96.02  & 56.03   & 61.03\\  
$\gamma^{(5)}$ & 61.03   & 51.02  & 51.02  & 71.02 & 51.02  & 61.03   & 56.03\\  
$\gamma^{(6)}$ & 86.03   & 91.01  & 91.01  & 86.03 & 96.02  & 81.03   & 86.03\\  
$\gamma^{(7)}$ & 91.01   & 91.01  & 91.01  & 91.01 & 86.03  & 86.03   & 96.02\\  
$\gamma^{(8)}$ & 91.01   & 91.01  & 91.01  & 91.01 & 91.01  & 91.01   & 86.03\\  
$\gamma^{(9)}$ & 96.02   & 96.02  & 96.02  & 96.02 & 96.02  & 96.02   & 91.01\\  
$\gamma^{(10)}$ & 91.01  & 91.01  & 91.01  & 91.01 & 91.01  & 96.02   & 96.02\\  
$\gamma^{(11)}$ & 91.01  & 91.01  & 91.01  & 91.01 & 91.01  & 91.01   & 81.03\\  
$\gamma^{(12)}$ & 66.01  & 66.01  & 66.01  & 66.01 & 61.03  & 61.03   & 71.02\\  
$\gamma^{(13)}$ & 91.01  & 91.01  & 91.01  & 91.01 & 91.01  & 91.01   & 86.03\\  
$\gamma^{(14)}$ & 00.00  & 00.00  & 00.00  & 00.00 & 00.00  & 00.00   & 00.00\\  
\midrule
Pruned~$(\%)$ & 84.02 & 84.12 & 84.17 & \textbf{84.46} & 76.13 & 82.59 & 76.99 \\  
\bottomrule
\end{tabular}
\label{tab:phi_retraining}
\end{table*}

\subsection{Procedure for Upper Pruning Percentage Limit of Layers}
Across all the experiments, when using our set of operating constraints to define $\gamma$, we collect the performance of an SVM model across $c \in \{1,2,\dots,99\}$.

\subsection{Evaluation of Estimator}
\paragraph{Run-Time}
To compare the improvement offered by our hash-based ACMI estimator, we choose the Minimum Spanning Tree-based (MST) CMI estimator from MINT~\cite{ganesh2020mint} as the nearest competitive baseline.
In this experiment, we apply both estimators over the $9^{th}$ convolution layer of VGG16.
To ensure fair comparison, we use ACMI with $\varphi=1$ as well as $\norm{\textrm{weights}}_2$ where weights are scaled to be between $[0,1]$ within each layer, use the grouping formulation introduced in MINT as well as a manual threshold $\delta$ on the ACMI values.
Here, we vary $G$ values for both the layer $l$ and $l+1$ (8 and 9) over 16, 32, 64, 128 and 256.
We use an average run-time from 10 trials, except for groups 128 and 256 for the MST-based estimator for which we use 2 trials.
Most importantly, we set 200 samples per class which results in a total of 2000 samples of activations used by the estimators.

\paragraph{Selection of $\varphi$}
We implement a number of possible functions and evaluate them over the CIFAR10-VGG16 experimental setup.
The exact hyper-parameters used to obtain ACMI values and obtain the final test accuracy are provided in Tables~\ref{tab:base_retraining}, and \ref{tab:phi_retraining}.
We maintain $G=64$ throughout these experiments.
The retraining performances are based on the highest Pruned ($\%$) at which the model has a test accuracy that matches or exceeds $93.43\%$ (from MINT).

\subsection{Large Scale Comparison}
The basic setup to obtain the final results presented in Table 2 of the main paper is listed under Table~\ref{tab:base_retraining}.
The main differences in the pruning setup between these experiments and the ones listed under Estimator evaluation are, 1) we avoid using a separate $\delta$ parameter and instead prune layers up to $\gamma^{(l)}$, and 2) we use label smoothing~\cite{szegedy2016rethinking}.
Below, we list the $\gamma$ values obtained through our set of operating constraints used to define the upper pruning percentage limit for all layers in the DNN.

For VGG16, $\gamma$ values from convolution layer 1 to the final linear layer are 0, 0, 0.8599, 0.9799, 0.9799, 0.9799, 0.9799, 0.9799, 0.9799, 0.9699, 0.9499, 0.8399, 0.9099, 0.

For ResNet56, $\gamma$ values from convolution layer 1 to the final linear layer are 0, 0, 0, 0, 0, 0, 0, 0, 0, 0, 0, 0, 0, 0, 0, 0, 0, 0, 0, 0, 0, 0, 0, 0, 0, 0, 0.699, 0.4794, 0.1591, 0.539, 0.3691, 0.9794, 0.089, 0.7392, 0.2695, 0.7294, 0.8896, 0.8398, 0.6699, 0.9699, 0.8698, 0.899, 0.2399, 0.9499, 0.3498, 0.899, 0.7199, 0.8898, 0.9199, 0.9599, 0.9699, 0.9799, 0.9799, 0.9799, 0.

For ResNet50, $\gamma$ values from convolution layer 1 to the final linear layer are
0.0, 0.0, 0.0, 0.0, 0.0, 0.0, 0.0, 0.0, 0.0, 0.0, 0.0, 0.0, 75.99, 65.99, 67.99, 96.99, 52.99, 82.99, 66.99, 78.99, 78.99, 51.99, 16.99, 67.99, 63.99, 76.99, 87.99, 66.00, 87.99, 92.99, 82.99, 73.99, 69.99, 93.99, 52.99, 95.99, 60.99, 78.99, 84.99, 57.00, 84.99, 80.99, 85.99, 34.99, 68.99, 94.00, 80.00, 87.00, 70.99, 79.99, 91.99, 98.00, 0.0, 0.0.

\subsection{Sensitivity-based Pruning}
\begin{table}[ht!]
\centering
\caption{Comparison of $\gamma$ values in CIFAR10-ResNet56 when sensitive filters are protected}
\begin{tabular}{@{}lcc@{}}
\toprule
              & w/o Sensitivity & with Sensitivity \\ \midrule
$\gamma^{(27)}$  &  0.699   & 0.699 \\
$\gamma^{(28)}$  &  0.4794  & 0.4394 \\
$\gamma^{(29)}$  &  0.1591  & 0.5507 \\
$\gamma^{(30)}$  &  0.5390  & 0.7041 \\
$\gamma^{(31)}$  &  0.3691  & 0.5117 \\
$\gamma^{(32)}$  &  0.9794  & 0.1796 \\
$\gamma^{(33)}$  &  0.089   & 0.3183 \\
$\gamma^{(34)}$  &  0.7392  & 0.6611 \\
$\gamma^{(35)}$  &  0.2695  & 0.4287 \\
$\gamma^{(36)}$  &  0.7294  & 0.8261 \\
$\gamma^{(37)}$  &  0.8896  & 0.7739 \\
$\gamma^{(38)}$  &  0.8398  & 0.8198 \\
$\gamma^{(39)}$  &  0.6699  & 0.799 \\
$\gamma^{(40)}$  &  0.9699  & 0.9299 \\
$\gamma^{(41)}$  &  0.8698  & 0.7927 \\
$\gamma^{(42)}$  &  0.899   & 0.8999 \\
$\gamma^{(43)}$  &  0.2399  & 0.2299 \\
$\gamma^{(44)}$  &  0.9499  & 0.8957 \\
$\gamma^{(45)}$  &  0.3498  & 0.5817 \\
$\gamma^{(46)}$  &  0.899   & 0.8898 \\
$\gamma^{(47)}$  &  0.7199  & 0.7099 \\
$\gamma^{(48)}$  &  0.8898  & 0.8759 \\
$\gamma^{(49)}$  &  0.9199  & 0.8813 \\
$\gamma^{(50)}$  &  0.9599  & 0.9599 \\
$\gamma^{(51)}$  &  0.9699  & 0.9699 \\
$\gamma^{(52)}$  &  0.9799  & 0.9699 \\
$\gamma^{(53)}$  &  0.9799  & 0.9799 \\
$\gamma^{(54)}$  &  0.9799  & 0.9799 \\
\midrule
Compression$(\%)$ & 68.59 & 68.96 \\
\bottomrule
\end{tabular}
\label{tab:sens_gamma}
\end{table}

When using sensitivity-based pruning for ResNet56, we observe both an increase and decrease in final $\gamma$ values used to achieve higher Pruned~($\%$) when compared to the case without sensitivity.
In Table~\ref{tab:sens_gamma} we highlight the difference in $\gamma$ values achieved in each case.
It is important to note that while $\gamma$ values represent the limit up to which layers should be pruned, in our implementation we obtain this point by selecting $\eta$ value just below the point which triggers the fail-safe.
Hence, layers with a skew in the distribution of $\eta$ values tend to be pruned more.
% we highlight convolution layers 4, 5, 8, 9 and 12 are the main layers where we can save sensitive filters and observe a gain in compression.
% In Table~\ref{tab:sens_gamma} we present the updated $\gamma$ values for these layers when sensitivity is added.
% These $\gamma$ values were obtained after applying our operating constraints on $c \in \{2,4,\dots,98\}$.

% use section* for acknowledgment
\section*{Acknowledgment}
This work has been partially supported (Madan Ravi Ganesh and Jason J. Corso) by a Google Faculty Research Award, and NIST 60NANB17D191 and (Salimeh Yasaei Sekeh) by NSF 1920908; the findings are those of the authors only and do not represent any position of these funding bodies.

% Can use something like this to put references on a page
% by themselves when using endfloat and the captionsoff option.
\ifCLASSOPTIONcaptionsoff
  \newpage
\fi

\bibliographystyle{IEEEtran}
\bibliography{egbib}

% that's all folks
\end{document}